\documentclass[10pt,twocolumn,letterpaper]{article}

\usepackage[pagenumbers]{cvpr} %
\usepackage[accsupp]{axessibility} %

\usepackage[dvipsnames]{xcolor}

\definecolor{cvprblue}{rgb}{0.21,0.49,0.74}
\usepackage[pagebackref,breaklinks,colorlinks,citecolor=cvprblue]{hyperref}

\newlength\savewidth
\usepackage{times}
\usepackage{epsfig}
\usepackage{graphicx}
\usepackage{amsmath}
\usepackage{wrapfig,lipsum,booktabs}
\usepackage[dvipsnames]{xcolor}
\usepackage{mathtools} 
\usepackage{multirow}
\usepackage{graphicx}  
\usepackage{amssymb}
\usepackage{gensymb}
\usepackage{color}
\usepackage{float}

\newcommand{\cmark}{\ding{51}}
\newcommand{\xmark}{\ding{55}}
\newcommand{\ours}{CAT-Seg\xspace}
\usepackage{colortbl}
\usepackage{comment}

\newcommand{\hlrow}{\rowcolor{black!6}}

\makeatletter
\def\hlinewd#1{%
\noalign{\ifnum0=`}\fi\hrule \@height #1 \futurelet
\reserved@a\@xhline}

\usepackage{pifont}

\def\paperID{14262} %
\def\confName{CVPR}
\def\confYear{2024}

\title{CAT-Seg: Cost Aggregation for Open-Vocabulary Semantic Segmentation}

\author{
Seokju Cho\textsuperscript{1,}$^*$ \qquad Heeseong Shin\textsuperscript{1,}$^*$ \qquad Sunghwan Hong\textsuperscript{1} \\
Anurag Arnab\textsuperscript{2} \qquad Paul Hongsuck Seo\textsuperscript{1,\(\dagger\)} \qquad Seungryong Kim\textsuperscript{1,\(\dagger\)} \\[5pt]
\textsuperscript{1}Korea University \qquad \textsuperscript{2}Google Research\\
{\tt\small \{seokju\_cho, hsshin98, sung\_hwan, phseo, seungryong\_kim\}@korea.ac.kr}\\
{\tt\small aarnab@google.com}
}

\begin{document}
\maketitle
\begin{abstract}

Open-vocabulary semantic segmentation presents the challenge of labeling each pixel within an image based on a wide range of text descriptions. In this work, we introduce a novel cost-based approach to adapt vision-language foundation models, notably CLIP, for the intricate task of semantic segmentation.  Through aggregating the cosine similarity score, i.e., the cost volume between image and text embeddings, our method potently adapts CLIP for segmenting seen and unseen classes by fine-tuning its encoders, addressing the challenges faced by existing methods in handling unseen classes. Building upon this, we explore methods to effectively aggregate the cost volume considering its multi-modal nature of being established between image and text embeddings. Furthermore, we examine various methods for efficiently fine-tuning CLIP.

\end{abstract}
\let\thefootnote\relax\footnotetext{$^*$Equal contribution. $^\dagger$Corresponding authors.}

\section{Introduction}

Open-vocabulary semantic segmentation aims to assign each pixel in an image to a class label from an unbounded range, defined by text descriptions. To handle the challenge of associating an image with a wide variety of text descriptions, pre-trained vision-language foundation models, \eg, CLIP~\cite{radford2021learning} and ALIGN~\cite{jia2021scaling}, have drawn attention as they exerted strong open-vocabulary recognition capabilities achieved through training on extensive image-text datasets. Nonetheless, these foundation models primarily receive image-level supervision during training, which introduces a notable disparity when applying them to the pixel-level segmentation tasks~\cite{zhou2022extract}. 

To address this gap, recent works~\cite{ding2022decoupling,ghiasi2022scaling,xu2022simple, liang2022open, xu2023open, xu2023side, yu2023convolutions} have reformulated the task into a region-level problem by utilizing mask proposal generators. While this partially bridges the discrepancy between the pre-training and the downstream task, a discernible gap persists between the conceptualization of regions and the entire image for CLIP. 

\begin{figure}[t]
  \centering

     \subfloat[mIoU of \textbf{seen} classes]{%
\includegraphics[width=0.49\linewidth]{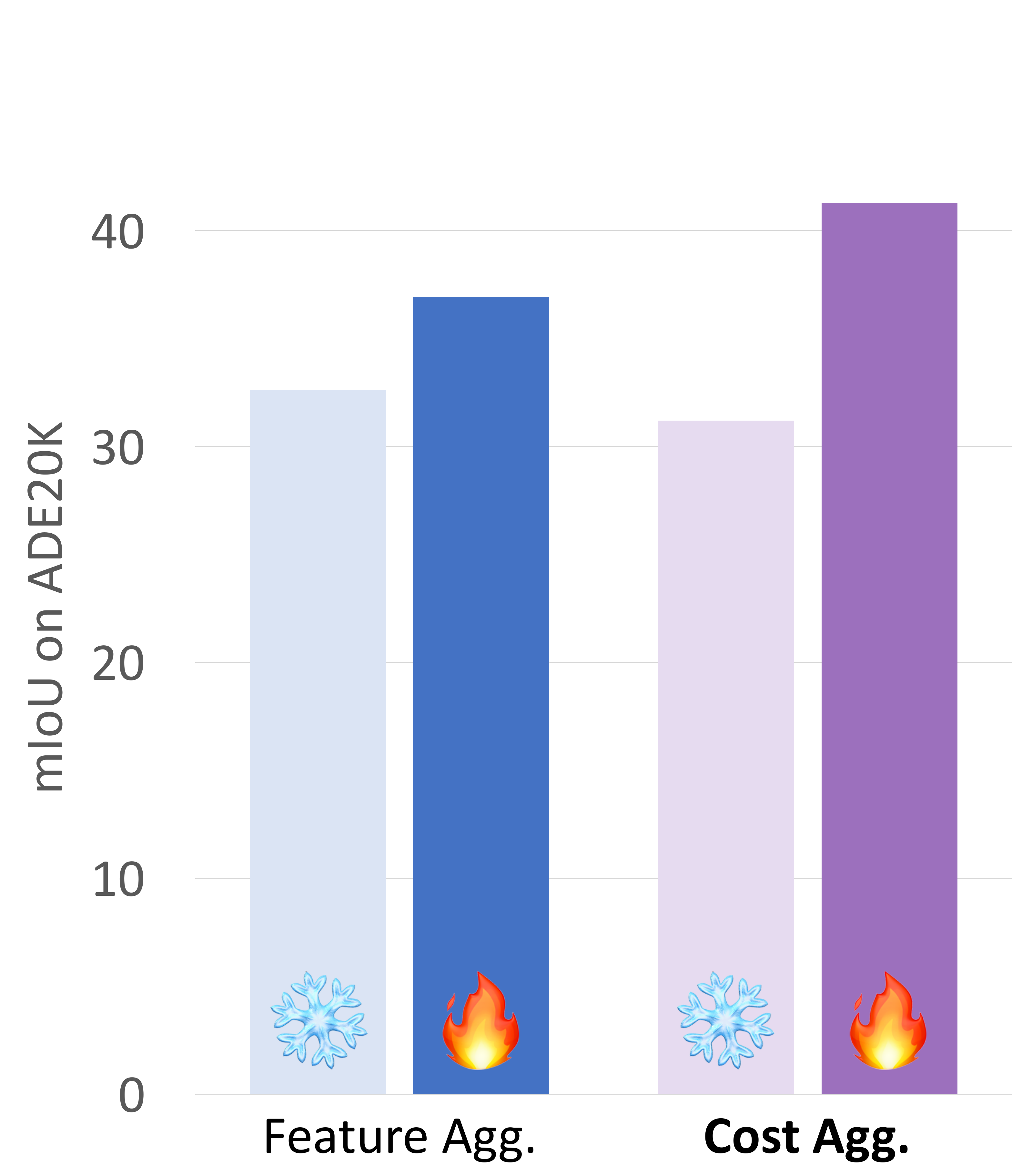}}\hfill
     \subfloat[mIoU of \textbf{unseen} classes]
{\includegraphics[width=0.49\linewidth]{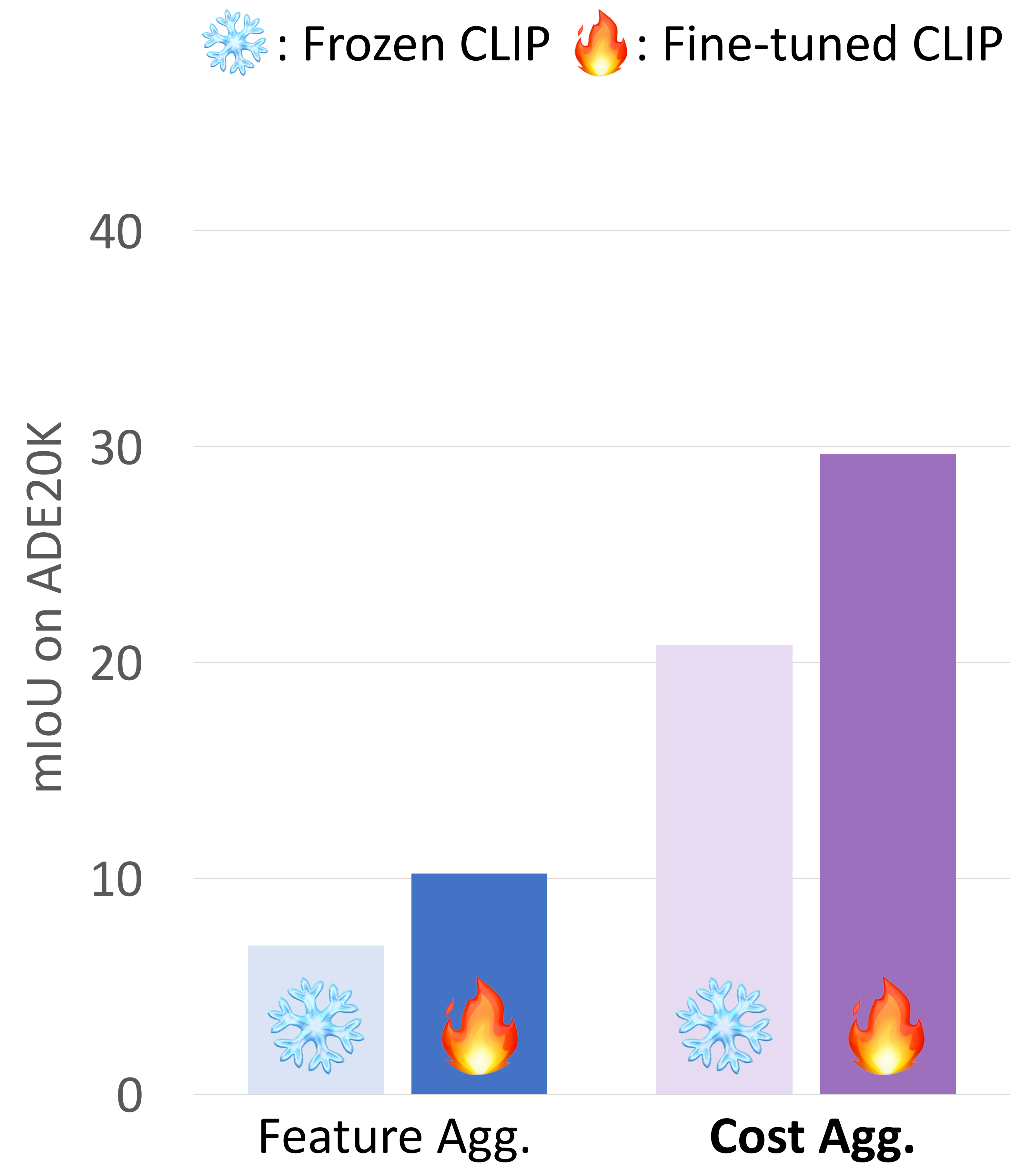}}\hfill

\vspace{-5pt}
\caption{\textbf{Comparison between feature and cost aggregation for open-vocabulary semantic segmentation task.} 
In contrast to feature aggregation suffering severe overfitting to seen classes, cost aggregation can generalize to unseen classes and achieve significant performance improvements upon fine-tuning of CLIP.%
}
\vspace{-10pt}
  \label{fig:motivation}
\end{figure}

In this work, we investigate methods to transfer the holistic understanding capability of images to the pixel-level task of segmentation. 
While a straightforward approach would be to fine-tune the encoders of CLIP, existing methods struggle in such attempt~\cite{zhou2022extract,yu2023convolutions,xu2023side} as they encounter significant overfitting problems to the seen classes.
This results in the misalignment of the joint embedding space for unseen classes, as the CLIP features undergo decoder modules for aggregating them into segmentation masks, hence losing their alignment.
Consequently, most methods~\cite{ding2022decoupling,ghiasi2022scaling,xu2022simple, liang2022open, xu2023open, xu2023side, yu2023convolutions} opt for freezing the encoders of CLIP instead, remaining the challenge underexplored.

In this regard, we extend the exploration of adapting CLIP for open-vocabulary semantic segmentation and introduce a novel cost-based framework. We propose to aggregate the cosine similarity between image and text embeddings of CLIP, \textit{i.e.}, the matching cost, drawing parallels to the visual correspondence literature~\cite{kendall2017end}. Surprisingly, we find that fine-tuning CLIP upon this framework effectively adapts CLIP to the downstream task of segmentation for both seen and unseen classes, as shown in Fig.~\ref{fig:motivation}. Noticing this, we delve into better aggregating the cost volume between image and text for segmentation.

\begin{figure}[t]
  \centering
    \subfloat[CLIP]
{\includegraphics[width=0.33\linewidth]{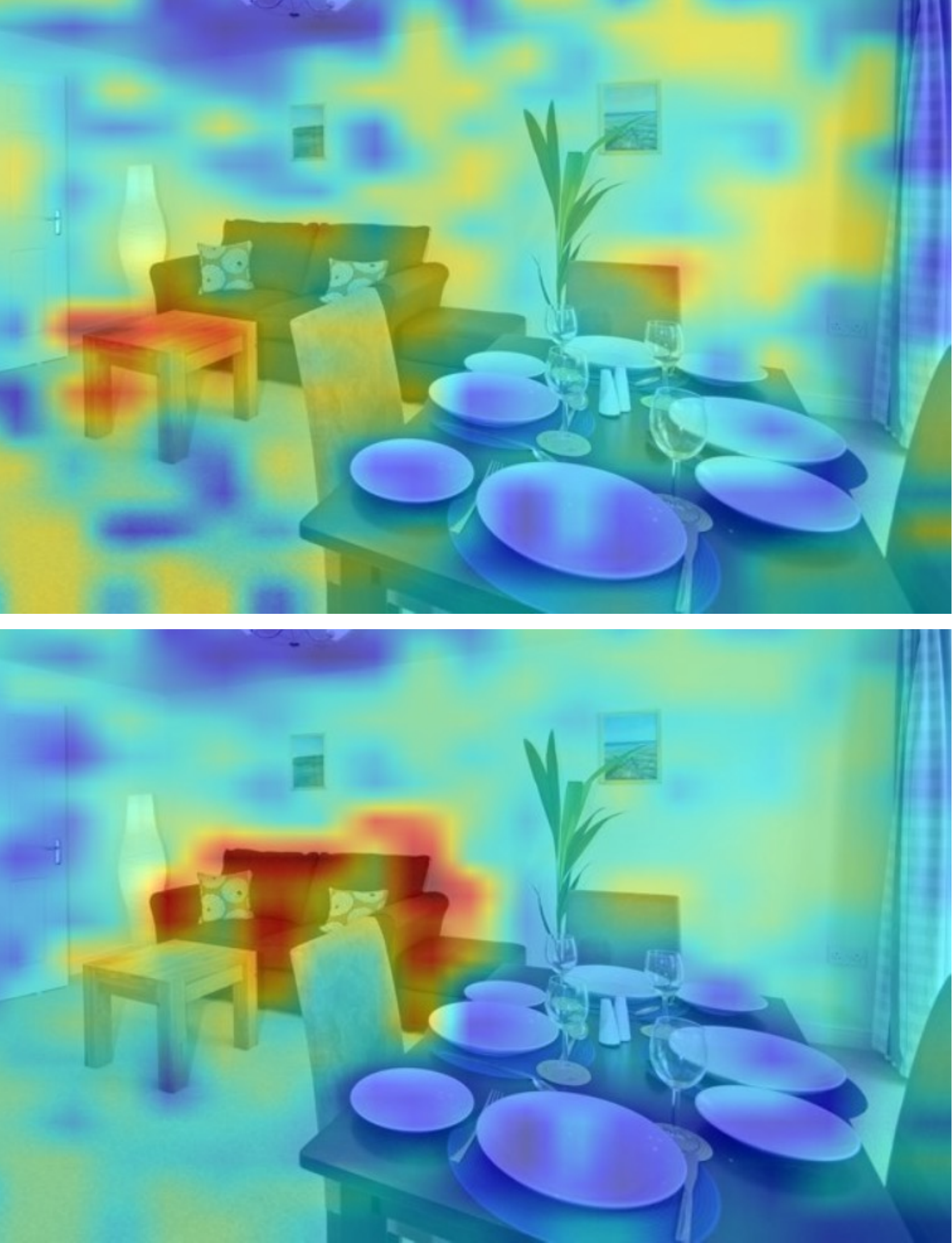}}\hfill
    \subfloat[Fine-tuned CLIP]
{\includegraphics[width=0.33\linewidth]{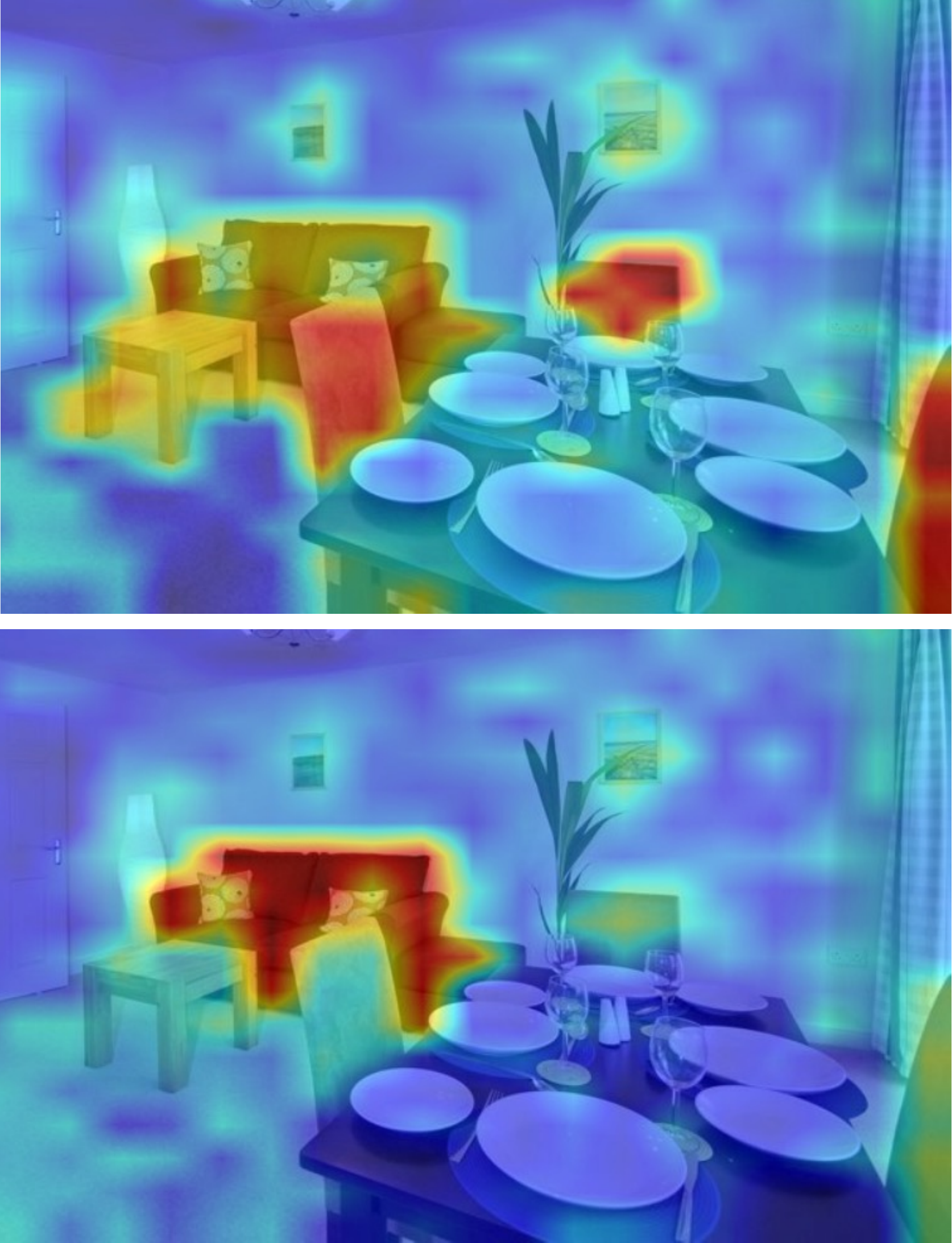}}\hfill
     \subfloat[Aggregated Cost]
 {\includegraphics[width=0.33\linewidth]{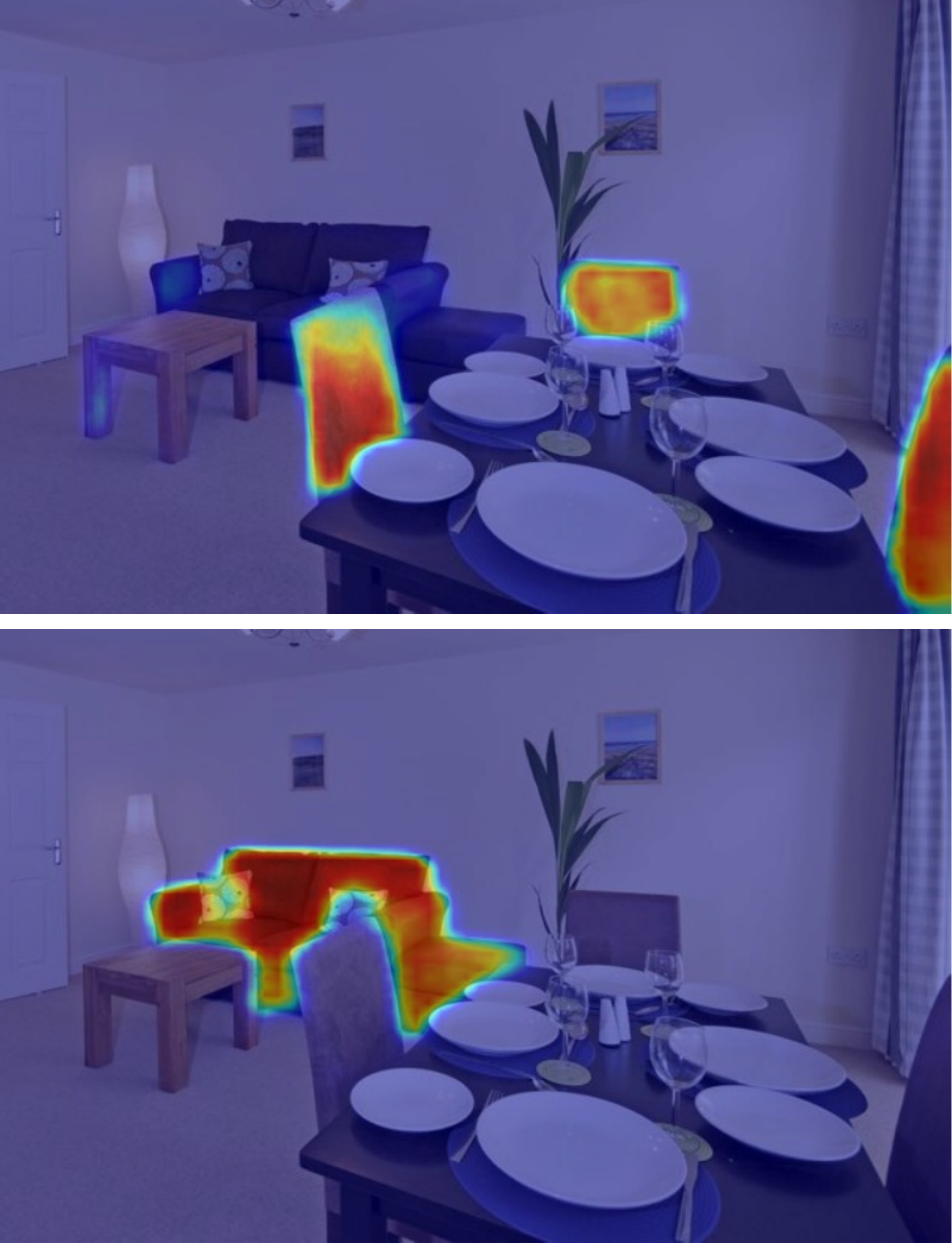}}\hfill\\
    
\vspace{-5pt}
\caption{\textbf{Visualization of the cost volume.} We visualize the raw cost volume obtained from frozen CLIP in (a) and fine-tuned CLIP in (b), and the aggregated cost in (c) through \ours. The top row correspond to the seen class ``chair" and the bottom row correspond to the unseen class ``sofa".} 
\label{fig:intuition}
\vspace{-10pt}
\end{figure}

Intuitively, the cost volume can be viewed as rough semantic masks grounded to their respective classes, as illustrated in Fig.~\ref{fig:intuition}. Subsequently, these rough masks can be further refined to obtain accurate predictions, being the cost aggregation process. In light of this, we aim to effectively aggregate the cost volume and configure the process into spatial and class aggregation, regarding its multi-modal nature from being established between image and text. Furthermore, by observing the effectiveness of fine-tuning CLIP for its adaptation to semantic segmentation, we explore various methods to facilitate this process efficiently.

We analyze our cost aggregation framework to be advantageous in two aspects for adapting CLIP to dense prediction: \textit{i}) the robustness of cost aggregation against overfitting, and \textit{ii}) the direct construction of the cost volume from image and text embeddings of CLIP. For cost aggregation, the aggregation layers operate upon similarity scores, preventing them from overfitting to the features~\cite{cai2020matching,song2021adastereo,liu2022graftnet}. Moreover, as opposed to existing methods where they often employ decoder layers upon the image embeddings of CLIP~\cite{zhou2022extract, yu2023convolutions}, we do not introduce additional layers that can potentially project the embeddings to a different embedding space. 

Our framework, dubbed \ours, combines our cost aggregation-based framework consisting of spatial and class aggregation, with our optimal approach for fine-tuning the encoders of CLIP. We achieve state-of-the-art results on every standard open-vocabulary benchmark with large margins, gaining +3.6 mIoU in A-847 and +8.1 mIoU in PC-459 compared to the recent state-of-the-art. Not only \ours it is effective, but is also efficient both for training and inference compared to region-text methods, being over $\times$3.7 faster for inference. Furthermore, even in the extreme scenario~\citep{blumenstiel2023mess} where the domain of the image and text description differs significantly from the training dataset, our model outperforms existing state-of-the-art methods with a large margin, paving the way for various domain-specific applications. 

We summarize our contribution as follows:
\begin{itemize}
  \item We propose a cost aggregation-based framework for open-vocabulary semantic segmentation, effectively adapting CLIP to the downstream task of segmentation by fine-tuning its encoders. 
  \item To aggregate the image-text cost volume, we consist of our framework with spatial and class aggregation to reason the multi-modal cost volume and explore various methods to enhance our cost aggregation framework.
  \item Our framework, named \ours, establishes state-of-the-art performance for standard open-vocabulary benchmarks, as well as for extreme case scenarios~\cite{blumenstiel2023mess}, demonstrating versatility and practicality.
\end{itemize}

\section{Related Work}
\paragraph{Open-vocabulary semantic segmentation.} Classical approaches to the task~\cite{zhao2017open, bucher2019zero, xian2019semantic} attempt to learn visual embeddings that align with pre-defined text embeddings~\cite{miller1998wordnet, mikolov2013efficient}. However, the limited vocabulary of the words has been the major bottlenecks. To address this, LSeg~\cite{li2022language} leveraged CLIP for learning pixel-level visual embeddings aligned with the text embeddings of CLIP. Alternatively, OpenSeg~\cite{ghiasi2022scaling} proposed to identify local regions within the image and correlate with the text embeddings with class-agnostic region proposals. Similarly, ZegFormer~\cite{ding2022decoupling} and ZSseg~\cite{xu2022simple} proposed two-stage frameworks for dealing with the task. Typically, they first learn to predict class-agnostic region proposals similar to~\cite{ghiasi2022scaling}, and feed them to CLIP for final predictions. To better recognize these regions, OVSeg~\cite{liang2022open} collects region-text pairs to fine-tune the CLIP encoder, while MaskCLIP~\cite{ding2022open} leverages the self-attention map from CLIP to refine the region proposals. Alternatively, ODISE~\cite{xu2023open} leverages pre-trained Stable Diffusion~\cite{Rombach_2022_CVPR} model for generating high-quality class-agnostic masks. However, these region-to-text matching methods~\cite{ding2022decoupling,ghiasi2022scaling,xu2022simple, liang2022open, xu2023open, xu2023side, yu2023convolutions} require a region generator, which is trained on a limited scale of annotated datasets.

More recently, ZegCLIP~\cite{zhou2022zegclip} and SAN~\cite{xu2023side} proposed one-stage frameworks, where they attempt to leverage the embeddings from CLIP to predict masks instead of having class-agnostic mask generators parallel to CLIP. Although these methods can better leverage the pre-trained knowledge from CLIP, they introduce learnable tokens or adapter layers to the CLIP image encoder, which can be only trained on the seen classes. FC-CLIP~\cite{yu2023convolutions} implements CLIP as the visual backbone for the segmentation model but opts for a frozen image encoder as they find fine-tuning the image encoder hinders performance for unseen classes. In contrast, we refrain from adding external layers to CLIP and achieve fine-tuning of its encoders by aggregating the cost volume, which is obtained solely from the embeddings of CLIP.

\vspace{-10pt}
\paragraph{Fine-tuning vision-language models.} Along with the advance of large-scale vision-language models, \eg CLIP, numerous attempts have been made to adapt CLIP to various downstream tasks~\cite{wortsman2022robust}. CoOp~\cite{zhou2022learning} and CoCoOp~\cite{zhou2022conditional} learn prompt tokens instead of optimizing the full model. Another stream of work is CLIP-Adapter~\cite{gao2023clip} and TIP-Adapter~\cite{zhang2021tip}, where they aggregate the image and text embeddings from CLIP through adapter layers instead of tuning the encoder itself. 
However, such methods mainly focus on few-shot settings rather than zero-shot evaluation. We explore end-to-end fine-tuning of CLIP for zero-shot pixel-level prediction, which has failed in numerous attempts~\cite{zhou2022extract, xu2023side, yu2023convolutions}.

\vspace{-10pt}
\paragraph{Cost aggregation.} Cost aggregation~\cite{kendall2017end, chang2018pyramid, guo2019group, yang2019hierarchical, song2021adastereo,hong2022neural,huang2022flowformer,cho2022cats++} is a popular technique adopted for the process of establishing correspondence between visually or semantically similar images~\cite{kendall2017end,guo2019group,yang2019hierarchical,cho2021cats,hong2022cost}
by reducing the impact of errors and inconsistencies in the matching process. A matching cost, an input to cost aggregation, is typically constructed between dense features extracted from a pair of images~\cite{rocco2017convolutional}, and often cosine-similarity~\cite{liu2022graftnet,rocco2017convolutional} is used. 
In this work, we view the cosine-similarity score between image and text embeddings of CLIP from the viewpoint of establishing the matching cost volume. Especially, we find the robustness of the cost aggregation layers to be favorable to open-vocabulary semantic segmentation, as these layers operate upon the similarity scores rather than the embeddings itself~\cite{song2021adastereo,liu2022graftnet}. However, our approach diverges from traditional methods as the cost volume obtained from CLIP is inherently multi-modal, originating from both image and text modalities. This contrasts with conventional cost aggregation techniques~\cite{kendall2017end,guo2019group,yang2019hierarchical,cho2021cats,hong2022cost}. Consequently, we explore methods to effectively aggregate the multi-modal cost volume.

\section{Methodology}

\begin{figure*}
    \centering
    \includegraphics[width=1.0\linewidth]{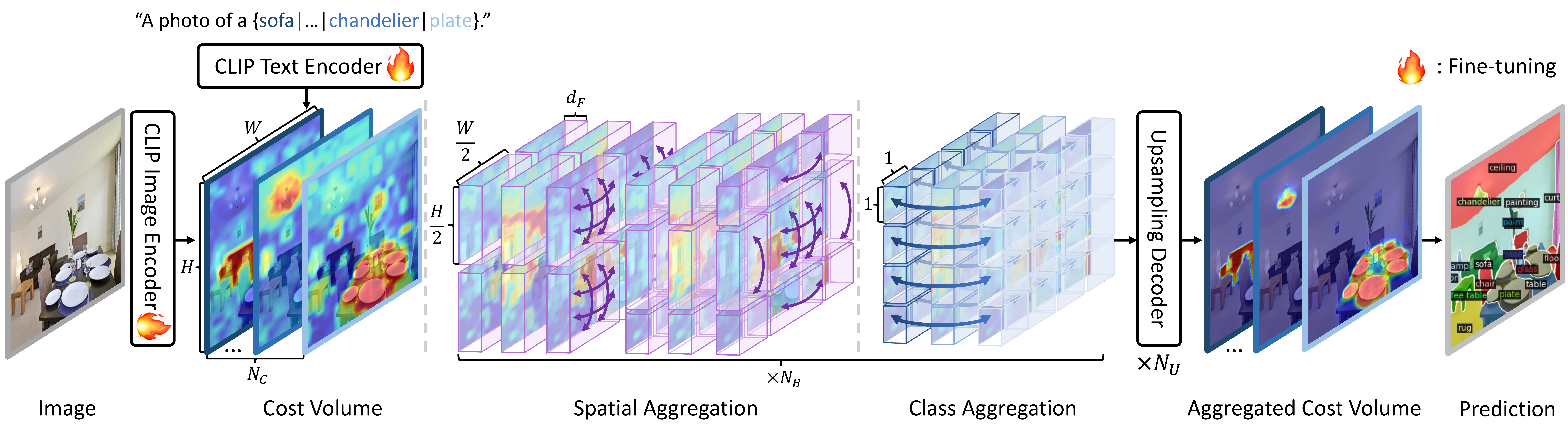}\hfill\\
    \vspace{-10pt}
    \caption{\textbf{Overview of \ours.} Our cost aggregation framework consists of spatial aggregation and class aggregation, followed by an upsampling decoder. Please refer to the supplementary material for a detailed illustration.}
    \label{fig:overall}\vspace{-10pt}
\end{figure*}

Given an image $I$ and a set of candidate class categories $\mathcal{C}=\{T(n)\} \text{ for } n=1,\dots ,N_{\mathcal{C}}$, where $T(n)$ denotes textual description of $n$-th category and $N_\mathcal{C}$ is the number of classes, open-vocabulary semantic segmentation assigns a class label for each pixel in image $I$.
Different from classical semantic segmentation tasks~\cite{long2015fully, he2017mask, zhou2022rethinking, he2019adaptive, jin2021mining, yu2020context, yuan2020object}, open-vocabulary segmentation is additionally challenged by varying $\mathcal{C}$, given as free-form text description.

In this section, we describe our cost-based approach for open-vocabulary semantic segmentation. In specific, we refine the cosine-similarity scores from image and text embedding of CLIP, as illustrated in Fig.~\ref{fig:intuition}. The process of refining the cosine-similarity scores, or cost aggregation~\cite{kendall2017end}, was initially developed for the image correspondence problem and specifically designed to process an image-to-image cost volume. Consequently, traditional cost aggregation methods leverage image-specific priors, such as the assumption of local smoothness of images~\cite{min2021convolutional, min2021hypercorrelation, lee2021deep} for aggregating the cost volume.

On the other hand, we aim to aggregate the image-to-text cost volume, hence need to consider the multi-modality of the cost volume and the respective characteristics of each modality. 
In this regard, as shown in Fig.~\ref{fig:overall}, we break down the aggregation stage into two separate modules, \ie, spatial and class aggregation, reasonably addressing the unique challenges presented by the task of open-vocabulary semantic segmentation. This includes aspects such as handling varying numbers of classes during inference and guaranteeing the permutation invariance between classes. Specifically, we perform spatial aggregation followed by class aggregation and alternate both aggregations. In the following section, we describe the cost aggregation process in detail, as well as introduce additional techniques for enhancing the cost aggregation framework.

\subsection{Cost Computation and Embedding}\label{sec:cost-computation}
Given an image $I$ and a set of classes $\mathcal{C}$, we extract the dense image embeddings $D^V = \Phi^V(I) \in \mathbb{R}^{(H \times W )\times d}$ and the text embeddings $D^L = \Phi^L(T) \in \mathbb{R}^{N_{\mathcal{C}}\times d}$, where $\Phi^V(\cdot)$ and  $\Phi^L(\cdot)$, denotes the image and text encoders of CLIP respectively. For extracting dense CLIP image embeddings, we follow the method described in \cite{zhou2022extract}, wherein we modify the last attention layer of the image encoder to eliminate the pooling effect.
We use the image and text embeddings $D^V(i)$ and $D^L(n)$, where $i$ denotes 2D spatial positions of the image embedding and $n$ denotes an index for a class, to compute a cost volume $C\in\mathbb{R}^{(H\times W)\times N_\mathcal{C}}$ by cosine similarity~\cite{rocco2017convolutional}. 
Formally, this is defined as:
\begin{equation}
  {C}(i,n)=
    \frac{D^V(i)\cdot D^L(n)}{\|D^V(i)\|\|D^L(n)\|}.
\end{equation}
To enhance the processing of cost in high dimensional feature space, we feed the cost volume to a single convolution layer that processes each cost slice $C(:, n)\in \mathbb{R}^{(H\times W)\times 1}$ independently to obtain initial cost volume embedding $F\in \mathbb{R}^{(H\times W)\times N_\mathcal{C} \times d_F}$, where $d_F$ is the cost embedding dimension, as shown in Fig.~\ref{fig:overall}. 

\subsection{Spatial Cost Aggregation} \label{sec:spatial-aggregation}
For spatial aggregation, we aim to consider the characteristics of images within the image-text cost volume, such as spatial smoothness within the image. Specifically, we apply spatial aggregation for each class, respectively. Considering that we pursue the holistic understanding of images of CLIP to effectively transfer to segmentation, we adopt Transformer~\cite{vaswani2017attention,liu2021swin} over CNNs for its global~\cite{vaswani2017attention} or semi-global~\cite{liu2021swin,hong2022cost} receptive fields. In practice, we employ Swin Transformer~\cite{liu2021swin} for computational efficiency. We define this process as follows:
\begin{equation}
    {F}'(:, n) = \mathcal{T}^\mathrm{sa}(F(:, n)), 
    \label{eq:spatial-aggregation}
\end{equation}
where $F(:,n)\in \mathbb{R}^{(H\times W)\times d_F}$, and $\mathcal{T}^\mathrm{sa}(\cdot)$ denotes a pair of two consecutive Swin transformer block for spatial aggregation, where the first block features self-attention within a local window, followed by the second block with self-attention within shifted window. Note that we treat $d_F$ as channel dimensions for each token, and attention is computed within individual classes separately. Intuitively, we can roughly relate the process of spatial aggregation to the bottom row of Fig.~\ref{fig:intuition}, where the cost volume for ``sofa" is well-refined after aggregation, and the noise in the background region is suppressed.

\subsection{Class Cost Aggregation} \label{sec:class-aggregation}
Subsequent to spatial aggregation, class aggregation is applied to consider the text modality, explicitly capturing relationships between different class categories. We also consider the unique challenges of open-vocabulary semantic segmentation of handling varying numbers of categories $\mathcal{C}$ while being invariant to their ordering. To address these challenges, we employ a Transformer~\cite{vaswani2017attention} layer without position embedding for aggregation, as this can achieve both of the aforementioned criteria. This process is defined as: 
\begin{align}
    {F}''(i, :) = \mathcal{T}^\mathrm{ca}({F}'(i, :)),
    \label{eq:class-aggregation}
\end{align}
where $F'(i,:)\in \mathbb{R}^{N_{\mathcal{C}}\times d_F}$, and $\mathcal{T}^\mathrm{ca}(\cdot)$ denotes a transformer block for class aggregation. 
In contrast to spatial aggregation, we instead employ a linear transformer~\cite{katharopoulos2020transformers} as we do not need to consider spatial structure of the input tokens in this aggregation, as well as benefitting from the linear computational complexity with respect to the number of the tokens. The class aggregation process can be related to the top row of Fig.~\ref{fig:intuition}, where the aggregated cost volume depicts its prediction to only chairs and excluding the sofa, as both classes are given together for reasoning.

\subsection{\ours Framework} \label{upsampling}
Upon the aggregated cost volume through spatial and class aggregation, we further enhance our methodology by incorporating an upsampling and aggregation process to derive semantic segmentation predictions. Additionally, drawing insights from state-of-the-art cost aggregation techniques~\cite{cho2021cats, cho2022cats++, hong2022cost}, we refine our cost aggregation strategy by leveraging guidance derived from the embeddings of CLIP. Finally, we examine various methods to fine-tune the encoders of CLIP, in pursuit of effectively, yet efficiently adapting CLIP for open-vocabulary semantic segmentation. Altogether, we introduce \textbf{C}ost \textbf{A}ggrega\textbf{T}ion approach for open-vocabulary semantic \textbf{Seg}mentation (\ours). 
We describe the upsampling decoder, embedding guidance, and our fine-tuning approach in detail in the subsequent sections. For detailed illustrations of the architecture for each component, please refer to the supplementary materials.

\vspace{-10pt}
\paragraph{Upsampling decoder.} 
Similar to FPN~\cite{lin2017feature}, we employ bilinear upsampling on the aggregated cost volume and concatenate it with the corresponding level of feature map extracted from CLIP, followed by a convolutional layer with a 3$\times$3 kernel of fixed size. We iterate this process $N_U$ times, generating a high-resolution output which is fed into the prediction head for final inference.
To extract the high-resolution feature map, we avoid using an additional feature backbone that would introduce a heavy computational burden. Instead, similarly to~\cite{li2022exploring}, we extract these maps from the middle layers of the CLIP image encoder. Specifically, we extract the feature map from the output of intermediate layers of CLIP ViT~\cite{dosovitskiy2020image} and then upsample them using a single learnable transposed convolution layer. This approach allows us to efficiently leverage the well-learned representations of CLIP for obtaining detailed predictions. For additional details, refer to the supplementary materials.

\vspace{-10pt}
\paragraph{Embedding guidance.}
As a means to enhance the cost aggregation process, we additionally leverage the embeddings $D_L\text{ and }D_V$ to provide spatial structure or contextual information of the inputs. Intuitively, we aim to guide the process with embeddings, based on the assumption that visually or semantically similar input tokens, e.g., color or category, have similar matching costs, inspired by cost volume filtering~\cite{hosni2012fast,sun2018pwc} in stereo matching literature~\cite{scharstein2002taxonomy}.
Accordingly, we redefine Eq.~\ref{eq:spatial-aggregation} and Eq.~\ref{eq:class-aggregation} as:  
\begin{equation}
\begin{split}
    &{F}'(:, n) = \mathcal{T}^\mathrm{sa}([F(:, n);\mathcal{P}^V(D^V)]),\\
    &{F}''(i,:) = \mathcal{T}^\mathrm{ca}([F'(i,:);\mathcal{P}^L(D^L)]),
\end{split}
\end{equation}
where $[\cdot]$ denotes concatenation, $\mathcal{P}^V$ and $\mathcal{P}^L$ denote linear projection layer, $D^V\in \mathbb{R}^{(H\times W)\times d}$, and $D^L\in \mathbb{R}^{N_{\mathcal{C}}\times d}$, where $d$ denotes the feature dimension. Notably, we only provide the embeddings to query and key as we find this is sufficient for embedding guidance. 

\vspace{-10pt}
\paragraph{Efficient fine-tuning of CLIP} 
While we aim to fully adapt CLIP to the downstream task through fine-tuning its image and text encoders, fine-tuning such foundation models can scale up to hundreds of millions of parameters, being computationally expensive and memory-intensive. On the other hand, freezing some of its layers, not only would be more efficient but also can help CLIP preserve its original embedding space, allowing it to be more robust to overfitting. To this end, we extensively investigate which layers should be frozen within CLIP~\cite{dosovitskiy2020image}, among examining various approaches for fine-tuning pre-trained models. We provide a detailed analysis of our exploration in Sec.~\ref{sec:ablation}.

\begin{table*}[!t]
    \begin{center}
   
    \resizebox{\textwidth}{!}{
    \begin{tabular}{l|cccc|cccccc}
    \toprule

        Model & VLM & Additional Backbone & Training Dataset & Additional Dataset & A-847 & PC-459 & A-150 & PC-59 & PAS-20 & $\textnormal{PAS-20}^b$\\
        \midrule\midrule
        SPNet~\citep{xian2019semantic}  & - & ResNet-101 & PASCAL VOC & \xmark & - & - & - & 24.3 & 18.3 & - \\
        ZS3Net~\citep{bucher2019zero}  & - & ResNet-101 & PASCAL VOC & \xmark  & - & - & - & 19.4 & 38.3 & - \\
        LSeg~\citep{li2022language} & CLIP ViT-B/32 & ResNet-101 & PASCAL VOC-15  & \xmark & - & - & - & - & 47.4 & - \\
        LSeg+~\citep{ghiasi2022scaling} & ALIGN & ResNet-101 & COCO-Stuff & \xmark  & 2.5 & 5.2 & 13.0 & 36.0 & - & 59.0 \\
        ZegFormer~\citep{ding2022decoupling} & CLIP ViT-B/16 & ResNet-101 & COCO-Stuff-156 & \xmark   & 4.9 & 9.1 & 16.9 & 42.8 & 86.2 & 62.7 \\
        ZegFormer{$\dagger$}~\citep{ding2022decoupling} & CLIP ViT-B/16 & ResNet-101 & COCO-Stuff & \xmark & 5.6 & 10.4 & 18.0 & 45.5 & 89.5 & \underline{65.5} \\
        ZSseg~\citep{xu2022simple} & CLIP ViT-B/16 & ResNet-101 & COCO-Stuff & \xmark   & 7.0 & - & 20.5 & 47.7 & 88.4 & - \\
        OpenSeg~\citep{ghiasi2022scaling} & ALIGN & ResNet-101 & COCO Panoptic & \cmark  &  4.4 & 7.9 & 17.5 & 40.1 & - & 63.8 \\
        OVSeg~\citep{liang2022open} & CLIP ViT-B/16 & ResNet-101c & COCO-Stuff & \cmark  &  7.1 & 11.0 & 24.8 & 53.3 & 92.6 & - \\
        ZegCLIP~\citep{zhou2022zegclip} & CLIP ViT-B/16 & - & COCO-Stuff-156 & \xmark  & -&-&-&41.2& 93.6 & - \\
        SAN~\citep{xu2023side} & CLIP ViT-B/16 & - & COCO-Stuff & \xmark  & \underline{10.1} & \underline{12.6} & \underline{27.5} & \underline{53.8} & \underline{94.0} & - \\
        \hlrow 
        & & & & &  \textbf{12.0} &\textbf{19.0} &\textbf{31.8} &\textbf{57.5} & \textbf{94.6} & \textbf{77.3} \\
        \hlrow\multirow{-2}{*}{\ours (ours)} & \multirow{-2}{*}{CLIP ViT-B/16} & \multirow{-2}{*}{-} & \multirow{-2}{*}{COCO-Stuff} & \multirow{-2}{*}{\xmark}  & \textcolor{ForestGreen}{(+1.9)} & \textcolor{ForestGreen}{(+6.4)} & \textcolor{ForestGreen}{(+4.3)} & \textcolor{ForestGreen}{(+3.7)} & \textcolor{ForestGreen}{(+0.6)} & \textcolor{ForestGreen}{(+11.8)} \\
        \midrule
        LSeg~\citep{li2022language} & CLIP ViT-B/32 & ViT-L/16 & PASCAL VOC-15 & \xmark & - & - & - & - & 52.3 & - \\
        OpenSeg~\citep{ghiasi2022scaling} & ALIGN & Eff-B7 & COCO Panoptic & \cmark & 8.1 & 11.5 & 26.4 & 44.8 & - & \underline{70.2}\\
        OVSeg~\citep{liang2022open} & CLIP ViT-L/14 & Swin-B & COCO-Stuff & \cmark & 9.0 & 12.4 & 29.6 & 55.7 & 94.5 & - \\
        SAN~\citep{xu2023side} & CLIP ViT-L/14 & - & COCO-Stuff & \xmark & \underline{12.4} & \underline{15.7} & \underline{32.1} & \underline{57.7} & \underline{94.6} & - \\
        ODISE~\citep{xu2023open} & CLIP ViT-L/14 & Stable Diffusion & COCO-Stuff & \xmark & 11.1 & 14.5 & 29.9 & 57.3 & - & - \\
        \hlrow 
        &  & & & & \textbf{16.0} & \textbf{23.8} & \textbf{37.9} & \textbf{63.3} & \textbf{97.0} & \textbf{82.5}\\
        \hlrow\multirow{-2}{*}{\ours (ours)} & \multirow{-2}{*}{CLIP ViT-L/14} & \multirow{-2}{*}{-} & \multirow{-2}{*}{COCO-Stuff} & \multirow{-2}{*}{\xmark} & \textcolor{ForestGreen}{(+3.6)} & \textcolor{ForestGreen}{(+8.1)} & \textcolor{ForestGreen}{(+5.8)} & \textcolor{ForestGreen}{(+5.6)} & \textcolor{ForestGreen}{(+2.4)} & \textcolor{ForestGreen}{(+12.3)}\\

        \bottomrule
    \end{tabular}
    }
    \vspace{-5pt}
        \caption{\textbf{Quantitative evaluation on standard benchmarks.} The best-performing results are presented in bold, while the second-best results are underlined. Improvements over the second-best are highlighted in \textcolor{ForestGreen}{green}. $\dagger$: Re-implementation trained on full COCO-Stuff.}
    \vspace{-15pt}
    \label{tab:main_table}
    \end{center}

\end{table*}

\begin{table*}[!t]
    \begin{center}
    
    \resizebox{\textwidth}{!}{
    \begin{tabular}{l|cc|ccccc|c}
\toprule

Model & VLM & Additional Backbone & General & Earth Monit. & Medical Sciences & Engineering & Agri. and Biology & Mean \\
\midrule\midrule
\textit{Random (LB)} & - & - & \phantom{0}\textit{1.17} & \phantom{0}\textit{7.11} & \textit{29.51} & \textit{11.71} & \phantom{0}\textit{6.14} & \textit{10.27} \\
\textit{Best supervised (UB)} & - & - & \textit{48.62} & \textit{79.12} & \textit{89.49} & \textit{67.66} & \textit{81.94} & \textit{70.99} \\
\midrule
ZSSeg~\citep{xu2022simple} & CLIP ViT-B/16 & ResNet-101 & 19.98 & 17.98 & \underline{41.82} & 14.0\phantom{0} & 22.32 & 22.73 \\
ZegFormer~\citep{ding2022decoupling} & CLIP ViT-B/16 & ResNet-101 & 13.57 & 17.25 & 17.47 & 17.92 & \underline{25.78} & 17.57 \\
X-Decoder~\citep{zou2023generalized} & UniCL-T & Focal-T & 22.01 & 18.92 & 23.28 & 15.31 & 18.17 & 19.8\phantom{0} \\
OpenSeeD~\citep{zhang2023simple} & UniCL-B & Swin-T & 22.49 & 25.11 & \textbf{44.44} & 16.5\phantom{0} & 10.35 & 24.33 \\
SAN~\citep{xu2023side} & CLIP ViT-B/16 & - & \underline{29.35} & \underline{30.64} & 29.85 & \textbf{23.58} & 15.07 & \underline{26.74} \\

\hlrow & & & \textbf{38.69} & \textbf{35.91} & 28.09 & \underline{20.34} & \textbf{32.57} & \textbf{31.96} \\
\hlrow\multirow{-2}{*}{\ours (ours)} & \multirow{-2}{*}{CLIP ViT-B/16} & \multirow{-2}{*}{-} & \textcolor{ForestGreen}{(+9.34)} & \textcolor{ForestGreen}{(+5.27)} & \color{gray}{(-16.35)} & \color{gray}{(-3.24)} & \textcolor{ForestGreen}{(+6.79)} & \textcolor{ForestGreen}{(+5.22)} \\
\midrule
OVSeg~\citep{liang2022open} & CLIP ViT-L/14 & Swin-B & 29.54 & 29.04 & \textbf{31.9\phantom{0}} & 14.16 & \underline{28.64} & 26.94 \\
SAN~\citep{xu2023side} & CLIP ViT-L/14 & - & \underline{36.18} & \underline{38.83} & \underline{30.27} & \underline{16.95} & 20.41 & \underline{30.06} \\
\hlrow & & & \textbf{44.69} & \textbf{39.99} & 24.70 & \textbf{20.20} & \textbf{38.61} & \textbf{34.70} \\
\hlrow\multirow{-2}{*}{\ours (ours)} & \multirow{-2}{*}{CLIP ViT-L/14} & \multirow{-2}{*}{-} & \textcolor{ForestGreen}{(+8.51)} & \textcolor{ForestGreen}{(+1.16)} & \color{gray}{(-7.2)} & \textcolor{ForestGreen}{(+3.25)} & \textcolor{ForestGreen}{(+9.97)} & \textcolor{ForestGreen}{(+4.64)} \\
        \bottomrule
    \end{tabular}
    }

    \vspace{-5pt}        
    \caption{\textbf{Quantitative evaluation on MESS~\citep{blumenstiel2023mess}.} MESS includes a wide range of domain-specific datasets, which pose significant challenges due to their substantial domain differences from the training dataset. We report the average score for each domain. Please refer to the supplementary material for the results of all 22 datasets. \textit{Random} is the result of uniform distributed prediction which represents the lower-bound, while \textit{Best supervised} represents the upper-bound performance for the datasets.}
    \label{tab:mess}
    \vspace{-20pt}
    \end{center}
\end{table*}

\section{Experiments}
\subsection{Datasets and Evaluation}
We train our model on the COCO-Stuff~\cite{caesar2018coco}, which has 118k densely annotated training images with 171 categories, following \cite{liang2022open}. We employ the mean Intersection-over-Union (mIoU) as the evaluation metric for all experiments. For the evaluation, we conducted experiments on two different sets of datasets~\cite{zhou2019semantic,everingham2009pascal,mottaghi2014role}: a commonly used in-domain datasets~\cite{ghiasi2022scaling}, and a multi-domain evaluation set~\cite{blumenstiel2023mess} containing domain-specific images and class labels. 

\vspace{-10pt}
\paragraph{Datasets for standard benchmarks.} 
For in-domain evaluation, we evaluate our model on ADE20K~\cite{zhou2019semantic}, PASCAL VOC~\cite{everingham2009pascal}, and PASCAL-Context~\cite{mottaghi2014role} datasets. ADE20K has 20k training and 2k validation images, with two sets of categories: A-150 with 150 frequent classes and A-847 with 847 classes~\cite{ding2022decoupling}. PASCAL-Context contains 5k training and validation images, with 459 classes in the full version (PC-459) and the most frequent 59 classes in the PC-59 version. PASCAL VOC has 20 object classes and a background class, with 1.5k training and validation images. We report PAS-20 using 20 object classes. We also report the score for PAS-$20^b$, which defines the ``background" as classes present in PC-59 but not in PAS-20, as in \citet{ghiasi2022scaling}.

\vspace{-10pt}
\paragraph{Datasets for multi-domain evaluation.}
We conducted a multi-domain evaluation on the MESS benchmark~\cite{blumenstiel2023mess}, specifically designed to stress-test the real-world applicability of open-vocabulary models with 22 datasets. The benchmark includes a wide range of domain-specific datasets from fields such as earth monitoring, medical sciences, engineering, agriculture, and biology. Additionally, the benchmark contains a diverse set of general domains, encompassing driving scenes, maritime scenes, paintings, and body parts. We report the average scores for each domain in the main text for brevity. For the complete results and details of the 22 datasets, please refer to the supplementary material.

\subsection{Implementation Details}
We train the CLIP image encoder and the cost aggregation module with per-pixel binary cross-entropy loss. We set $d_F=128$, $N_B=2$, $N_U=2$ for all of our models. We implement our work using PyTorch~\cite{paszke2019pytorch} and Detectron2~\cite{wu2019detectron2}. AdamW~\cite{loshchilov2017decoupled} 
optimizer is used with a learning rate of $2\cdot10^{-4}$ for our model  and $2\cdot10^{-6}$ for the CLIP, with weight decay set to $10^{-4}$. The batch size is set to 4. We use 4 NVIDIA RTX 3090 GPUs for training. All of the models are trained for 80k iterations. 

\subsection{Main Results}
\begin{figure*}[t]
  \centering
    \subfloat[SAN]
{\includegraphics[width=0.1595\linewidth]{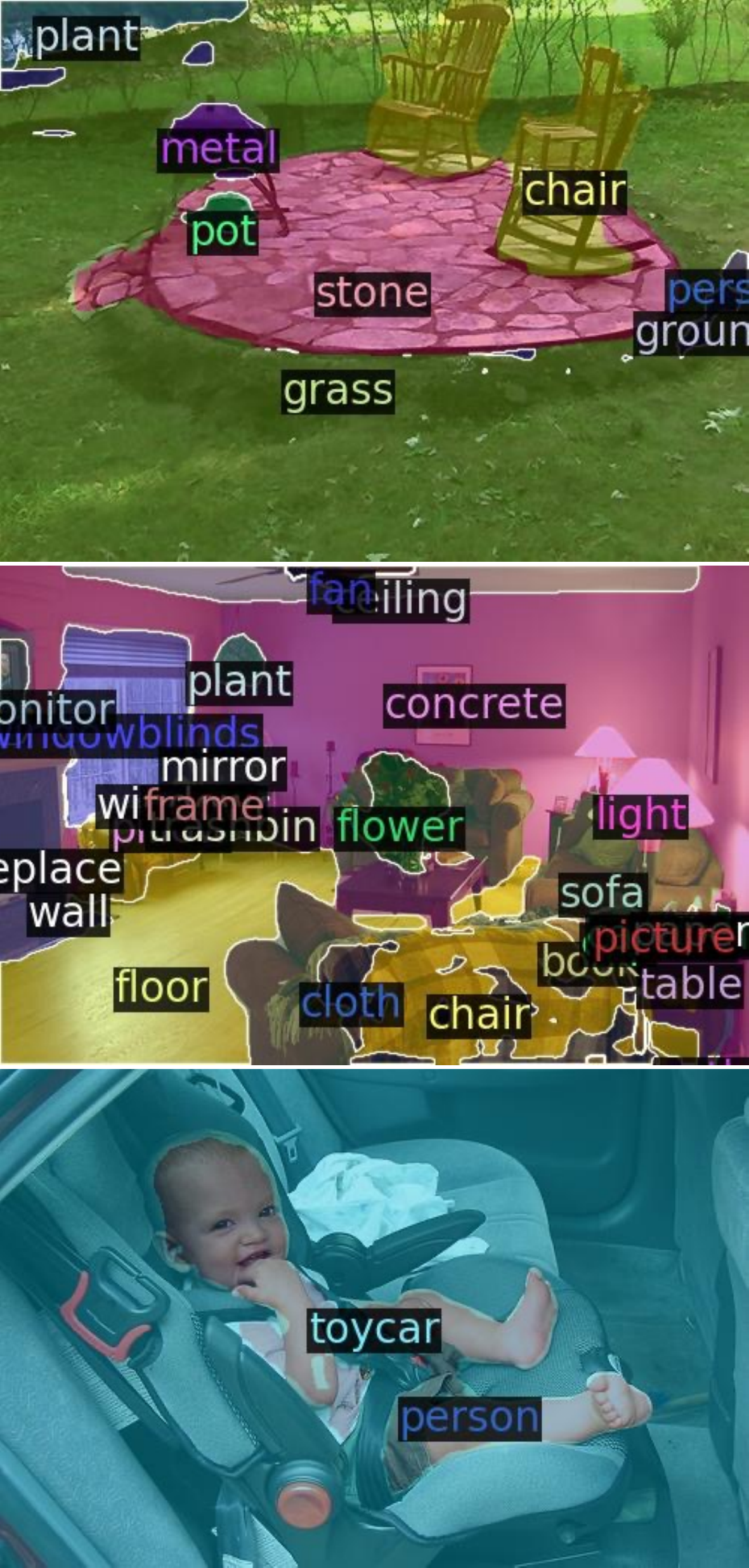}}\hfill
    \subfloat[\textbf{Ours}]
{\includegraphics[width=0.1595\linewidth]{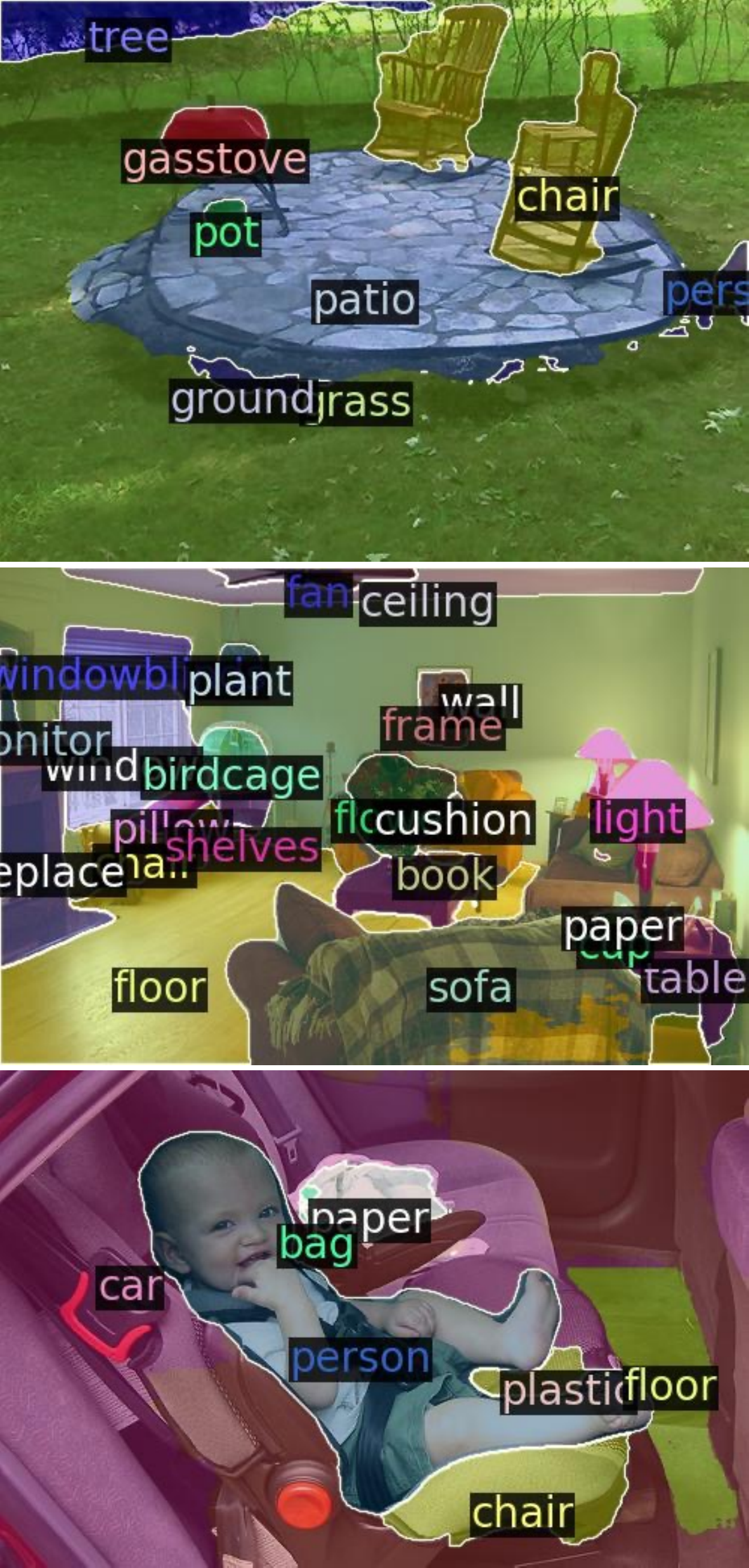}}\hfill
     \subfloat[GT]
 {\includegraphics[width=0.1595\linewidth]{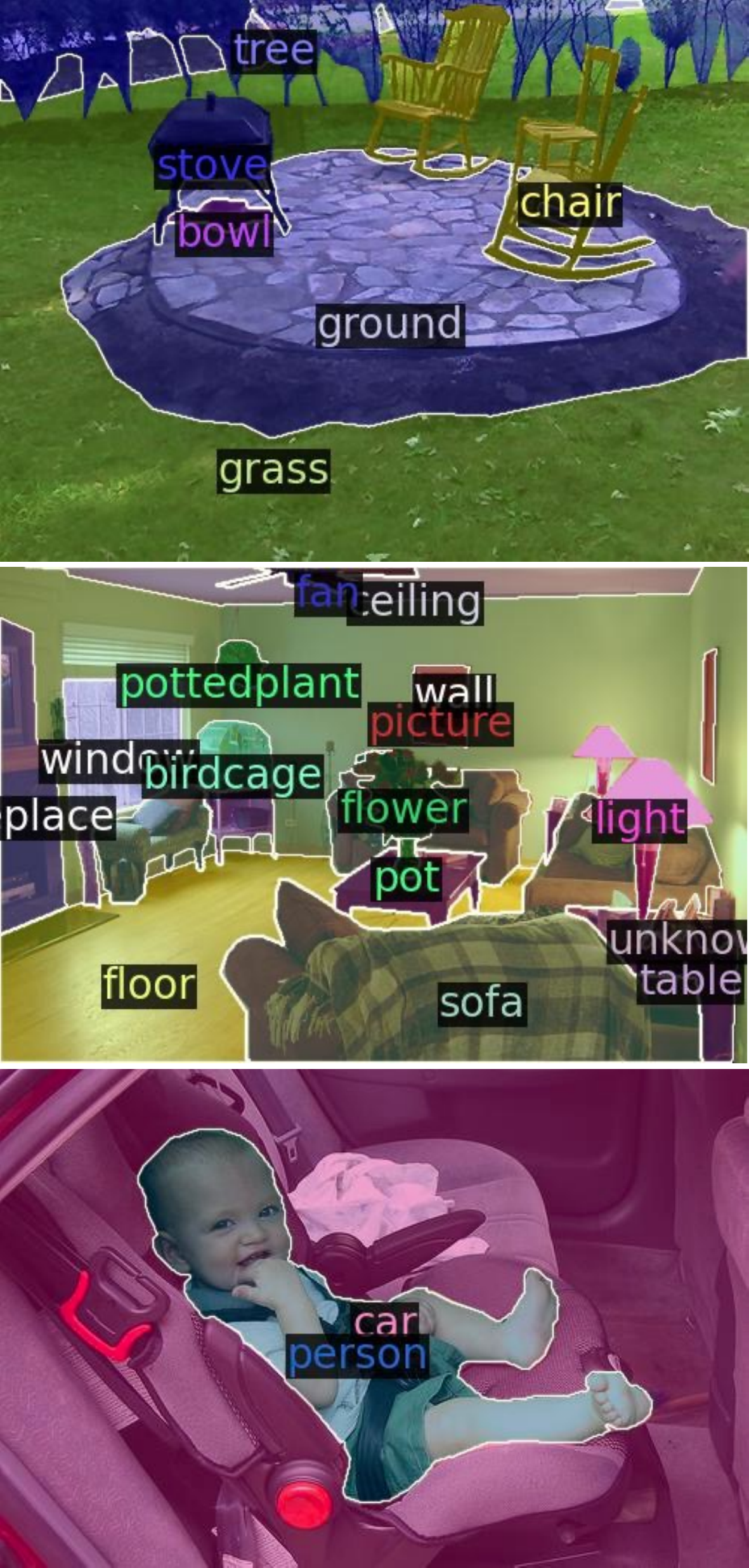}}\hfill
    \subfloat[SAN]
{\includegraphics[width=0.166\linewidth]{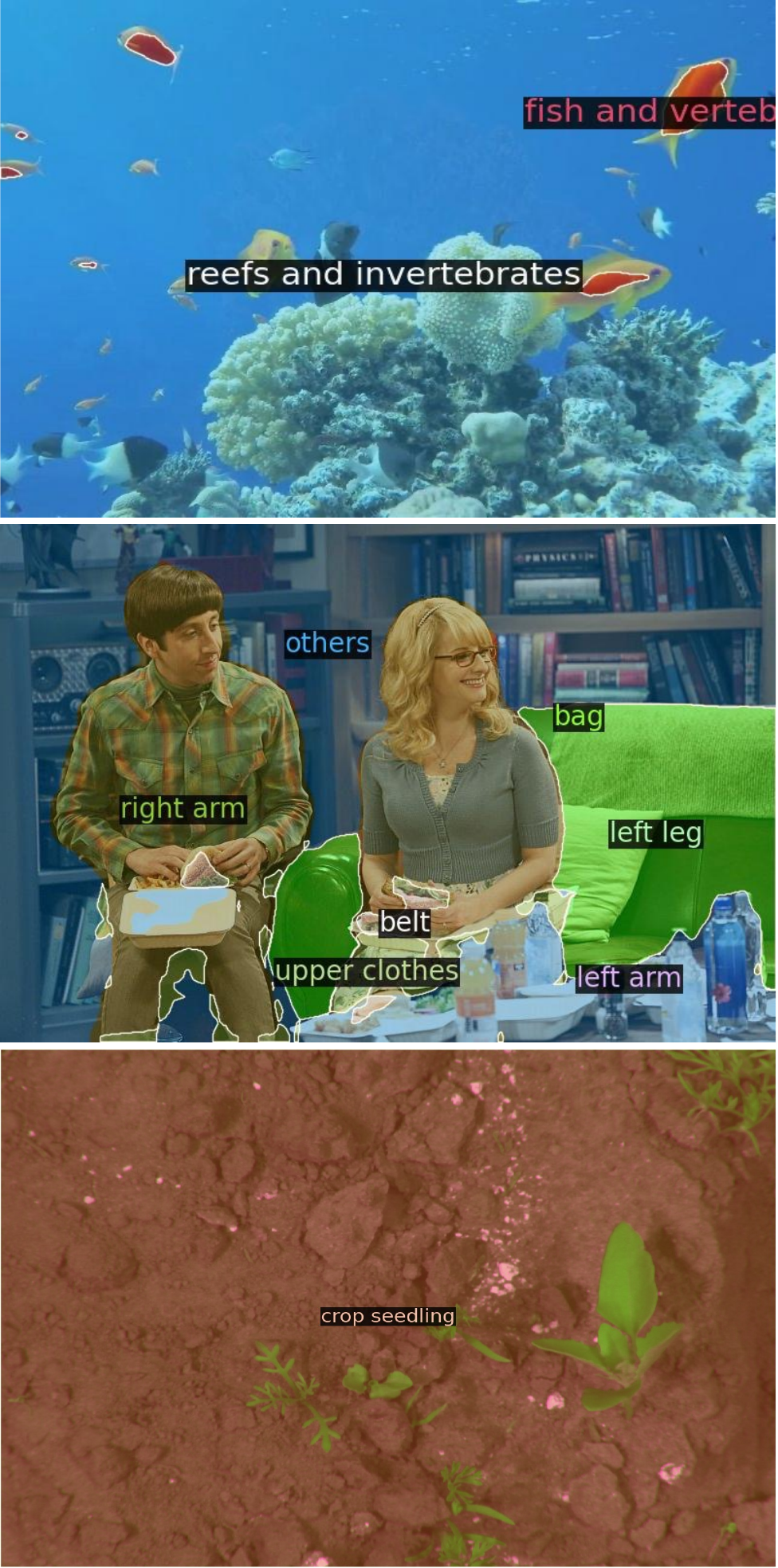}}\hfill
    \subfloat[\textbf{Ours}]
{\includegraphics[width=0.166\linewidth]{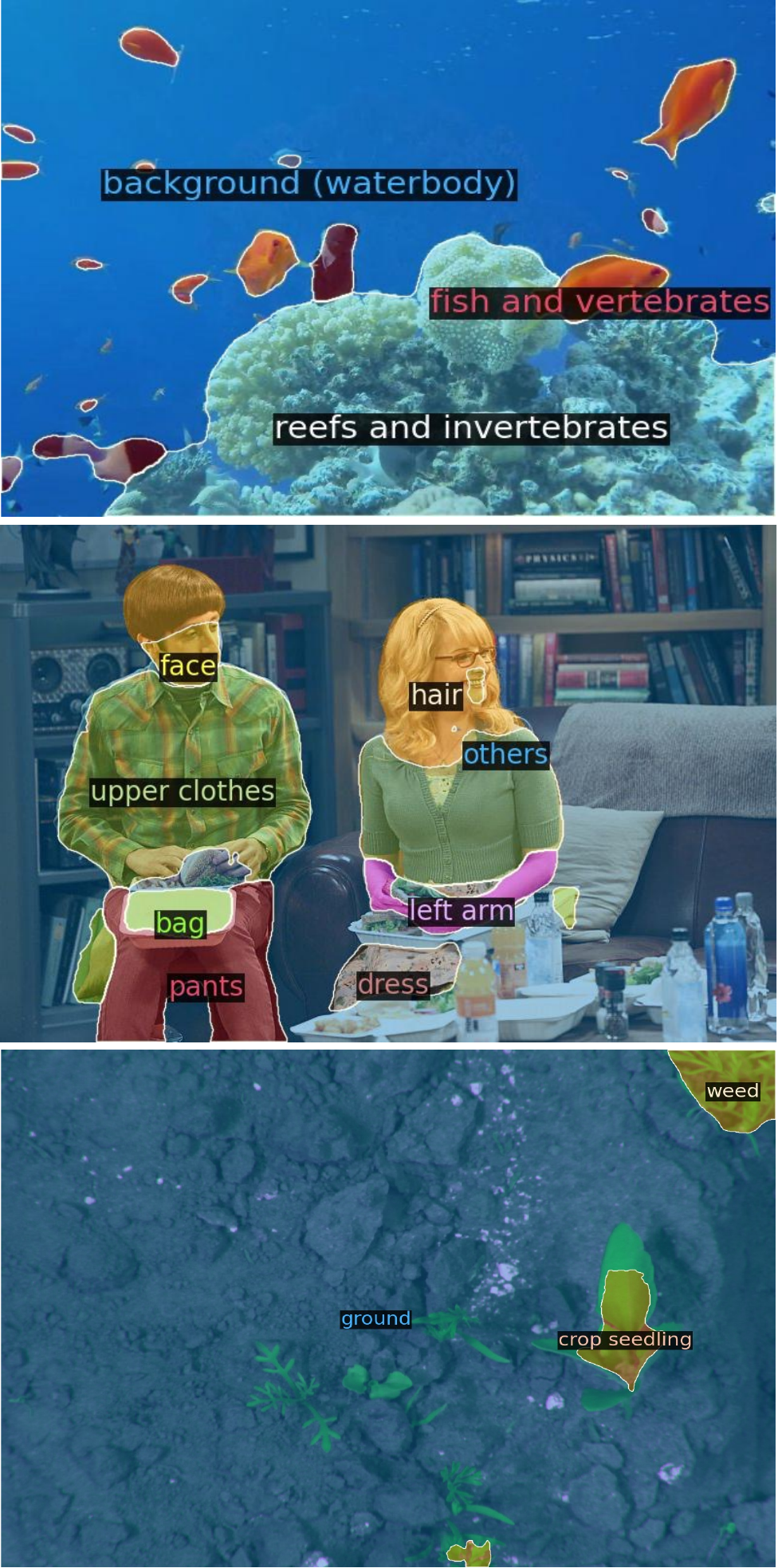}}\hfill
    \subfloat[GT]
{\includegraphics[width=0.166\linewidth]{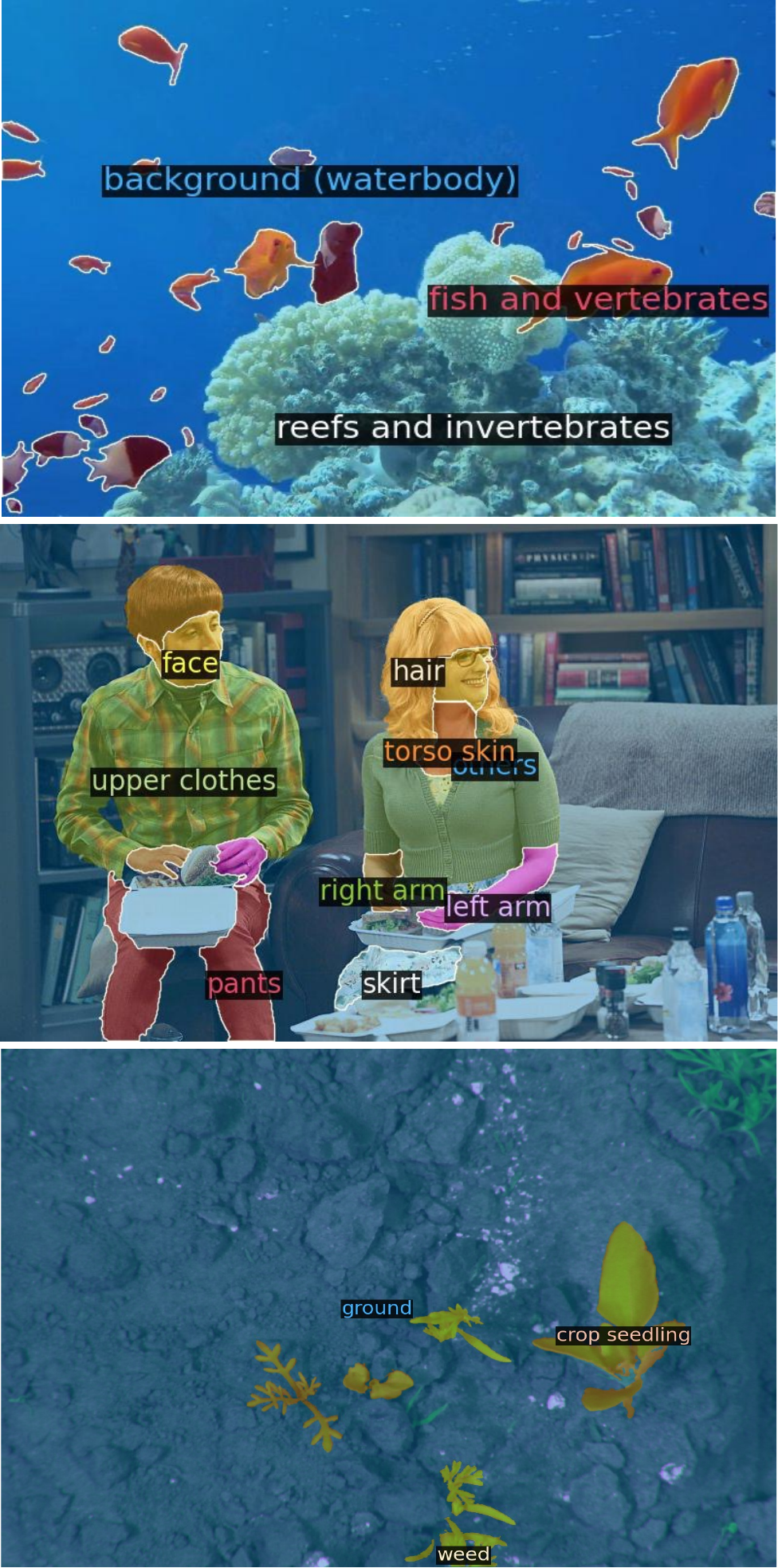}}\hfill
\\
\vspace{-10pt}
\caption{\textbf{Qualitative comparison to SAN~\citep{xu2023side}.} We visualize the results of PC-459 dataset in (a-c). For (d-f), we visualize the results from the MESS benchmark~\citep{blumenstiel2023mess} across three domains: underwater (top), human parts (middle), and agriculture (bottom).} 
\label{fig:qualitative}
\vspace{-10pt}
\end{figure*}

\paragraph{Results of standard benchmarks.}
The evaluation of standard open-vocabulary semantic segmentation benchmarks is shown in Table~\ref{tab:main_table}. Overall, our method significantly outperforms all competing methods, including those~\cite{ghiasi2022scaling,liang2022open} that leverage additional datasets~\cite{chen2015microsoft,pont2020connecting} for further performance improvements. To ensure a fair comparison, we categorize the models based on the scale of the vision-language models (VLMs) they employ. First, we present results for models that use VLMs of comparable scale to ViT-B/16~\cite{dosovitskiy2020image}, and our model surpasses all previous methods, even achieving performance that matches or surpasses those using the ViT-L/14 model as their VLM~\cite{xu2023side}.
For models employing the ViT-L/14 model as their VLM, our model demonstrates remarkable results, achieving a 16.0 mIoU in the challenging A-847 dataset and a 23.8 mIoU in PC-459. These results represent a 29\% and 52\% increase, respectively, compared to the previous state-of-the-art.
We also present qualitative results of PASCAL-Context with 459 categories in Fig.~\ref{fig:qualitative}, demonstrating the efficacy of our proposed approach in comparison to the current state-of-the-art methods~\cite{ding2022decoupling, xu2022simple,liang2022open}.

\begin{table}[t]
    \centering
    \resizebox{0.48\textwidth}{!}{%
    \begin{tabular}{cl|cccccc}
    \toprule
        & Methods & A-847 & PC-459 & A-150 & PC-59 & PAS-20 & $\textnormal{PAS-20}^b$
        \\
        \midrule\midrule
        \textbf{(I)} & Feature agg. + Freeze & 3.1 & 8.7 & 16.6 & 46.8 & 92.3 & 69.7\\
        \textbf{(II)} & Feature agg. + F.T. & 5.6 & {12.8} & {23.6} & \underline{58.1} & \underline{96.3} & \underline{77.7}\\
        \midrule
        \textbf{(III)} & Cost agg. + Freeze& \underline{10.0} & \underline{14.5} & \underline{26.0} & 46.9 & 94.2 & 65.1\\
        \hlrow\textbf{(IV)} & Cost agg. + F.T. & \textbf{14.7} & \textbf{23.2} & \textbf{35.3} & \textbf{60.3} & \textbf{96.7} & \textbf{78.9}\\
        \bottomrule
    \end{tabular}%
    }
    \vspace{-5pt}
    \caption{\textbf{Quantitative comparison between feature and cost aggregation.} Cost aggregation acts as an effective alternative to direct fine-tuning of CLIP image encoder. \textit{F.T.: Fine-Tuning.}
    }
    \label{tab:feature-vs-cost}
    \vspace{-15pt}

\end{table}

\vspace{-10pt}
\paragraph{Results of multi-domain evaluation.}
In Table~\ref{tab:mess}, we present the qualitative results obtained from the MESS benchmark~\cite{blumenstiel2023mess}. This benchmark assesses the real-world performance of a model across a wide range of domains. Notably, our model demonstrates a significant performance boost over other models, achieving the highest mean score. It particularly excels in the general domain as well as in agriculture and biology, showing its strong generalization ability. However, in the domains of medical sciences and engineering, the results exhibit inconsistencies with respect to the size of the VLM. Additionally, the scores for medical sciences are comparable to random predictions. We speculate that CLIP may have limited knowledge in these particular domains~\cite{radford2021learning}.

\subsection{Analysis and Ablation Study}\label{sec:ablation}

\paragraph{Comparison between feature and cost aggregation.} We provide quantitative and qualitative comparison of two aggregation baselines, feature aggregation, and cost aggregation, in Table~\ref{tab:feature-vs-cost}. For both of baseline architectures, we simply apply the upsampling decoder and note that both methods share most of the architecture, but differ in whether they aggregate the concatenated features or aggregate the cosine similarity between image and text embeddings of CLIP.
\begin{figure}[t]
  \centering
        \subfloat[Image]
{\includegraphics[width=0.495\linewidth]{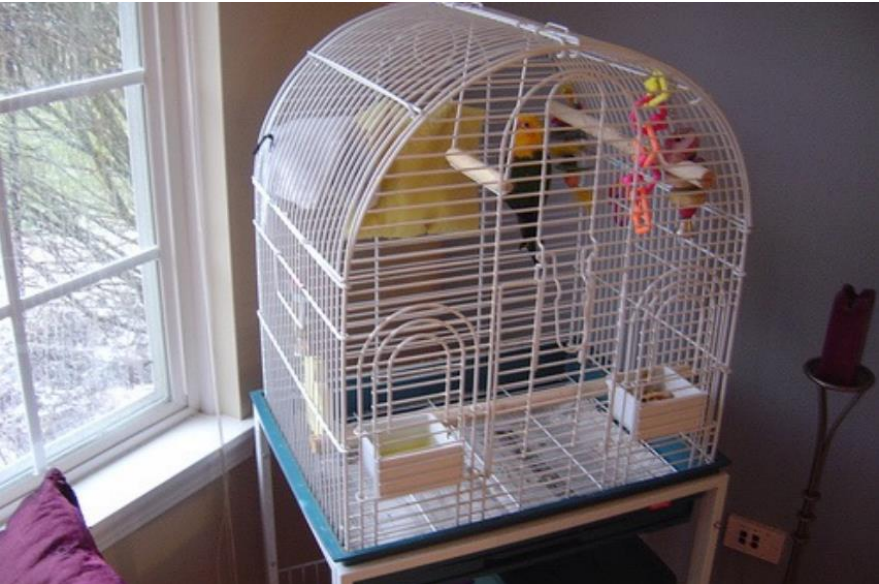}}\hfill
     \subfloat[Ground truth]
 {\includegraphics[width=0.495\linewidth]{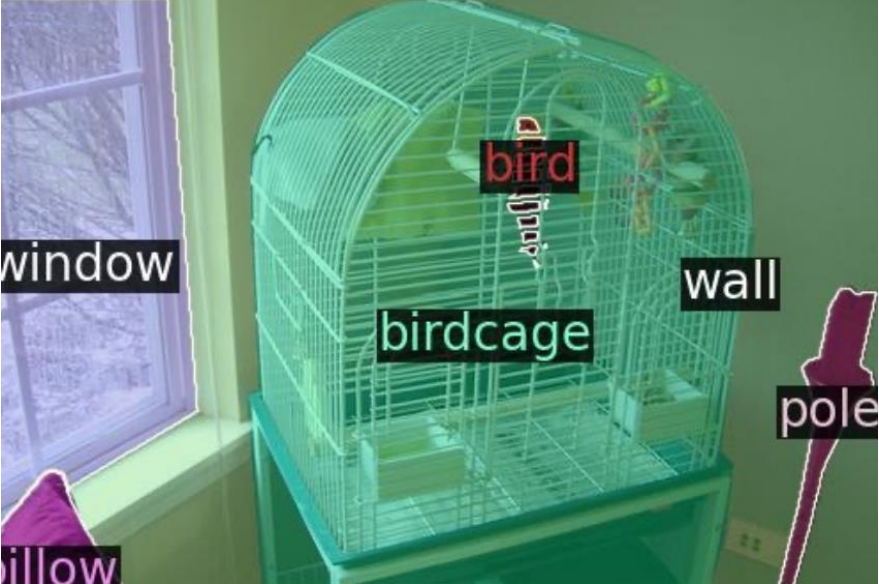}}\hfill\\
    \subfloat[Feature aggregation]
{\includegraphics[width=0.495\linewidth]{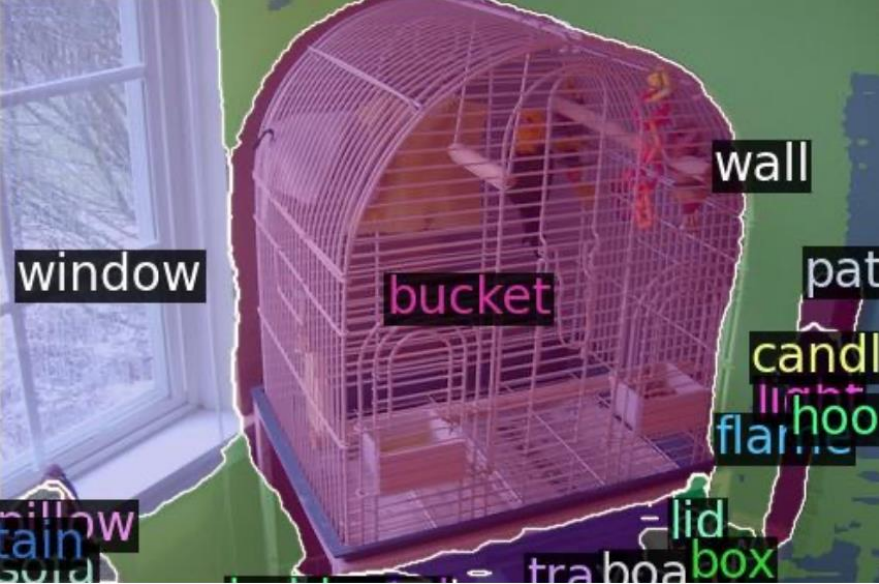}}\hfill
     \subfloat[Cost aggregation]
 {\includegraphics[width=0.495\linewidth]{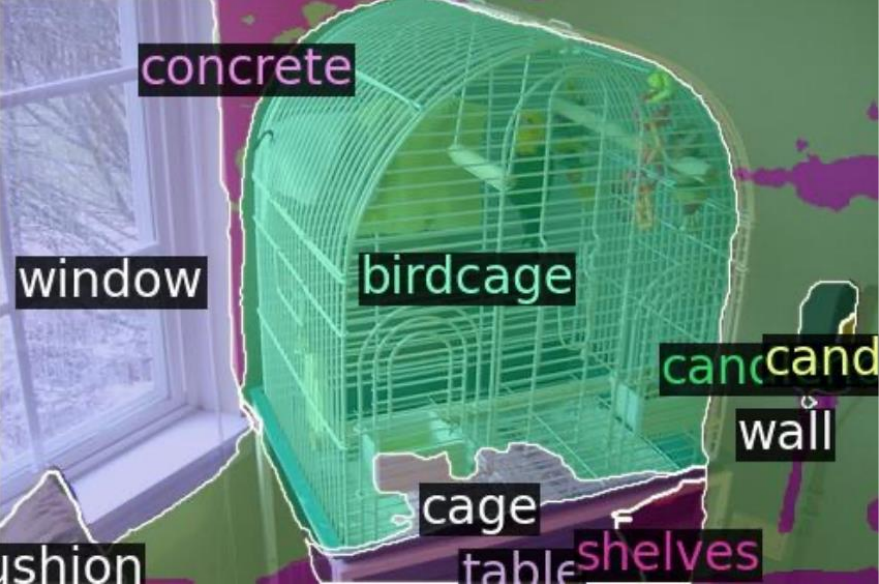}}\hfill\\

\vspace{-5pt}
\caption{\textbf{Qualitative comparison between feature and cost aggregation.} Our approach (d) successfully segments the previously unseen class, such as ``birdcage," whereas approach (c) fails.
}
\vspace{-10pt}
  \label{fig:feature_cost}
\end{figure}

For \textbf{(I)} and \textbf{(III)}, we freeze the encoders of CLIP and only optimize the upsampling decoder. Subsequently, in \textbf{(II)} and \textbf{(IV)}, we fine-tune the encoders of CLIP on top of \textbf{(I)} and \textbf{(III)}. Our results show that feature aggregation can benefit from fine-tuning, but the gain is only marginal. On the other hand, cost aggregation benefits significantly from fine-tuning, highlighting the effectiveness of cost aggregation for adapting CLIP to the task of segmentation. 

For the qualitative results in Fig.~\ref{fig:feature_cost}, we show the prediction results from \textbf{(II)} and \textbf{(IV)}. As seen in Fig.~\ref{fig:feature_cost}(c-d), we observe that feature aggregation shows overfitting to the seen class of ``bucket," while cost aggregation successfully identifies the unseen class ``birdcage." 

\begin{table}[t!]
\centering
\resizebox{\linewidth}{!}{
\begin{tabular}{ll|cccccc}
        \toprule
        &Components & A-847 & PC-459 & A-150 & PC-59 & PAS-20 & $\textnormal{PAS-20}^b$
         \\
        \midrule\midrule
        \textbf{(I)} & Feature Agg. & 5.6 & 12.8 & 23.6 & 58.1 & 96.3 & 77.7\\
        \midrule
        \textbf{(II)} & Cost Agg.  & 14.7& \underline{23.2}& 35.3& 60.3& \underline{96.7}&78.9\\
        \textbf{(III)} &\textbf{(II)} + Spatial agg.  & 14.9& 23.1& 35.9& 60.3& \underline{96.7}&79.5\\
        \textbf{(IV)} &\textbf{(II)} + Class agg.  & 14.7& 21.5& 36.6& 60.6& 95.5&80.5\\
        \textbf{(V)} &\textbf{(II)} + Spatial and Class agg. & \underline{15.5}& \underline{23.2}& \underline{37.0}& \underline{62.3}& \underline{96.7}&\underline{81.3}\\
        \hlrow\textbf{(VI)} &\textbf{(V)} + Embedding guidance  & \textbf{16.0} & \textbf{23.8}& \textbf{37.9}& \textbf{63.3}& \textbf{97.0}&\textbf{82.5}\\
        \bottomrule
\end{tabular}}
\vspace{-5pt}
\caption{\textbf{Ablation study for \ours.} We conduct ablation study by gradually adding components to the cost aggregation baseline.}
    \vspace{-10pt}
    \label{tab:ablation}
\end{table}

\begin{table}[t]
    \centering
    \resizebox{\linewidth}{!}{
   \begin{tabular}{l|cccccc}
        \toprule
        Methods & A-847 & PC-459 & A-150 & PC-59 & PAS-20 & $\textnormal{PAS-20}^b$
         \\
        \midrule\midrule
        \ours w/o upsampling decoder & \underline{9.9}& \underline{16.1}& \underline{28.4}& \underline{52.9}& \underline{93.2}& \underline{73.3}\\
        \hlrow\ours (ours) & \textbf{12.0} &\textbf{19.0} &\textbf{31.8} &\textbf{57.5} & \textbf{94.6} & \textbf{77.3}\\
        \bottomrule
\end{tabular}}
\vspace{-5pt}
        \caption{\textbf{Ablation study of upsampling decoder.} CLIP with ViT-B is used for ablation.}
    \label{tab:conv-decoder}
    \vspace{-10pt}
\end{table}

\vspace{-10pt}
\paragraph{Component analysis.}
Table~\ref{tab:ablation} shows the effectiveness of the main components within our architecture through quantitative results. 
First, we introduce the baseline models in \textbf{(I)} and \textbf{(II)}, identical to the fine-tuned baseline models from Table~\ref{tab:feature-vs-cost}.
We first add the proposed spatial and class aggregations to the cost aggregation baseline in \textbf{(III)} and \textbf{(IV)}, respectively. In \textbf{(V)}, we interleave the spatial and class aggregations. Lastly, we add the proposed embedding guidance to \textbf{(V)}, which becomes our final model.

As shown, we stress the gap between \textbf{(I)} and  \textbf{(II)}, which supports the findings presented in Fig.~\ref{fig:feature_cost}. Given that PAS-20 shares most of its classes with the training datasets\cite{xu2022simple}, the performance gap between \textbf{(I)} and \textbf{(II)} is minor. However, for challenging datasets such as A-847 or PC-459, the difference is notably significant, validating our cost aggregation framework for its generalizability.
We also highlight that as we incorporate the proposed spatial and class aggregation techniques, our approach \textbf{(V)} outperforms \textbf{(II)}, demonstrating the effectiveness of our design.
Finally, \textbf{(VI)} shows that our embedding guidance further improves performance across all the benchmarks.
Furthermore, we provide quantitative results of adopting the upsampling decoder in Table ~\ref{tab:conv-decoder}. The results show consistent improvements across all the benchmarks.

\begin{table}[!t]
\centering
\resizebox{\linewidth}{!}{
   \begin{tabular}{ll|cccccc|cc}
        \toprule
        &\multirow{2}{*}{Methods} & \multirow{2}{*}{A-847} & \multirow{2}{*}{PC-459} & \multirow{2}{*}{A-150} & \multirow{2}{*}{PC-59}& \multirow{2}{*}{PAS-20} & \multirow{2}{*}{$\textnormal{PAS-20}^b$} &\#param.  & Memory
         \\
         &&&&&&&&(M)&(GiB)
         \\
        \midrule\midrule
        \textbf{(I)} &Freeze & 10.4& 15.0& 31.8& 52.5& 92.2& 71.3& 5.8 & 20.0\\
        \textbf{(II)} &Prompt  & 8.8& 14.3 & 30.5& 55.8 & 93.2 & 74.7 & 7.0 & 20.9\\
        \textbf{(III)} &Full F.T.  & 13.6& 22.2& 34.0& 61.1& \textbf{97.3}& 79.7 & 393.2 & 26.8\\
        \textbf{(IV)} &Attn. F.T. & 15.7& \underline{23.7}& 37.1& \underline{63.1}& \underline{97.1}& 81.5 & 134.9 & 20.9\\
        \textbf{(V)} &QK F.T. & 15.3& 23.0& 36.3& 62.0& 95.9& 81.9 & 70.3 & 20.9\\
        \textbf{(VI)} &KV F.T. & \textbf{16.1}& \textbf{23.8}& \underline{37.6}& 62.4& 96.7& \underline{82.0} & 70.3 & 20.9\\
        \midrule
        \textbf{(VII)} & QV F.T. (Img.)  & 13.9& 22.8& 35.1& 62.0& 96.3& \underline{82.0} & 56.7 & 20.9\\
        \textbf{(VIII)} & QV F.T. (Txt.)  & 14.7& 22.2& 35.1& 60.0& 95.8& 80.3 & 19.9 & 20.0\\
        \hlrow \textbf{(IX)} & QV F.T. (Both) & \underline{16.0}& \textbf{23.8}& \textbf{37.9}& \textbf{63.3}& 97.0& \textbf{82.5} & 70.3 & 20.9\\
        \bottomrule       
\end{tabular}
}
    \vspace{-5pt}
\caption{\textbf{Analysis of fine-tuning methods for CLIP.} We additionally note the number of learnable parameters of CLIP and memory consumption during training. Our method not only outperforms full fine-tuning, but also requires smaller computation.}
\label{tab:finetuning-ablation}
\vspace{-10pt}
\end{table}

\begin{figure}[t]
  \centering
    \subfloat[CLIP]
{\includegraphics[width=0.4999\linewidth]{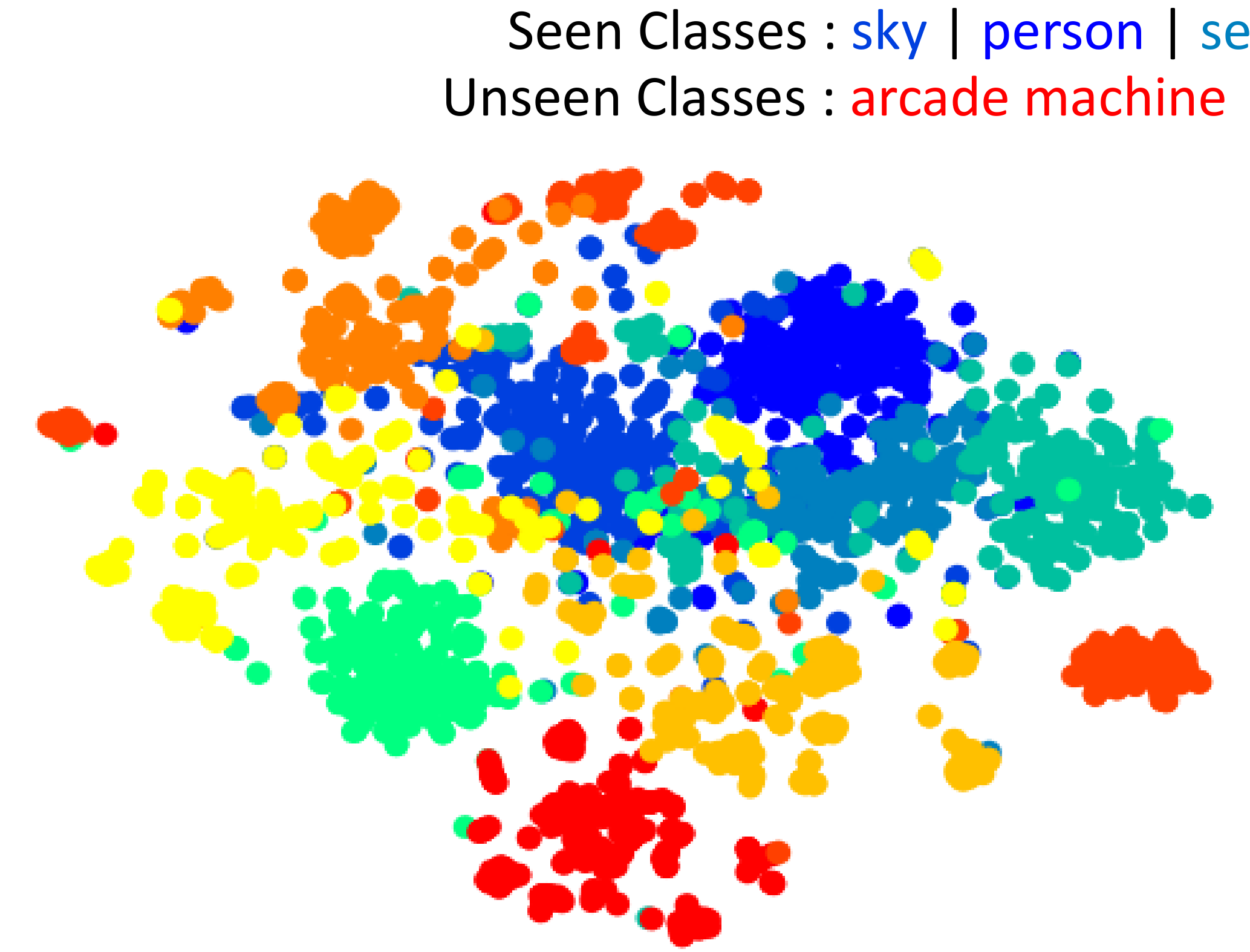}}\hfill
     \subfloat[Fine-tuned CLIP]
 {\includegraphics[width=0.4999\linewidth]{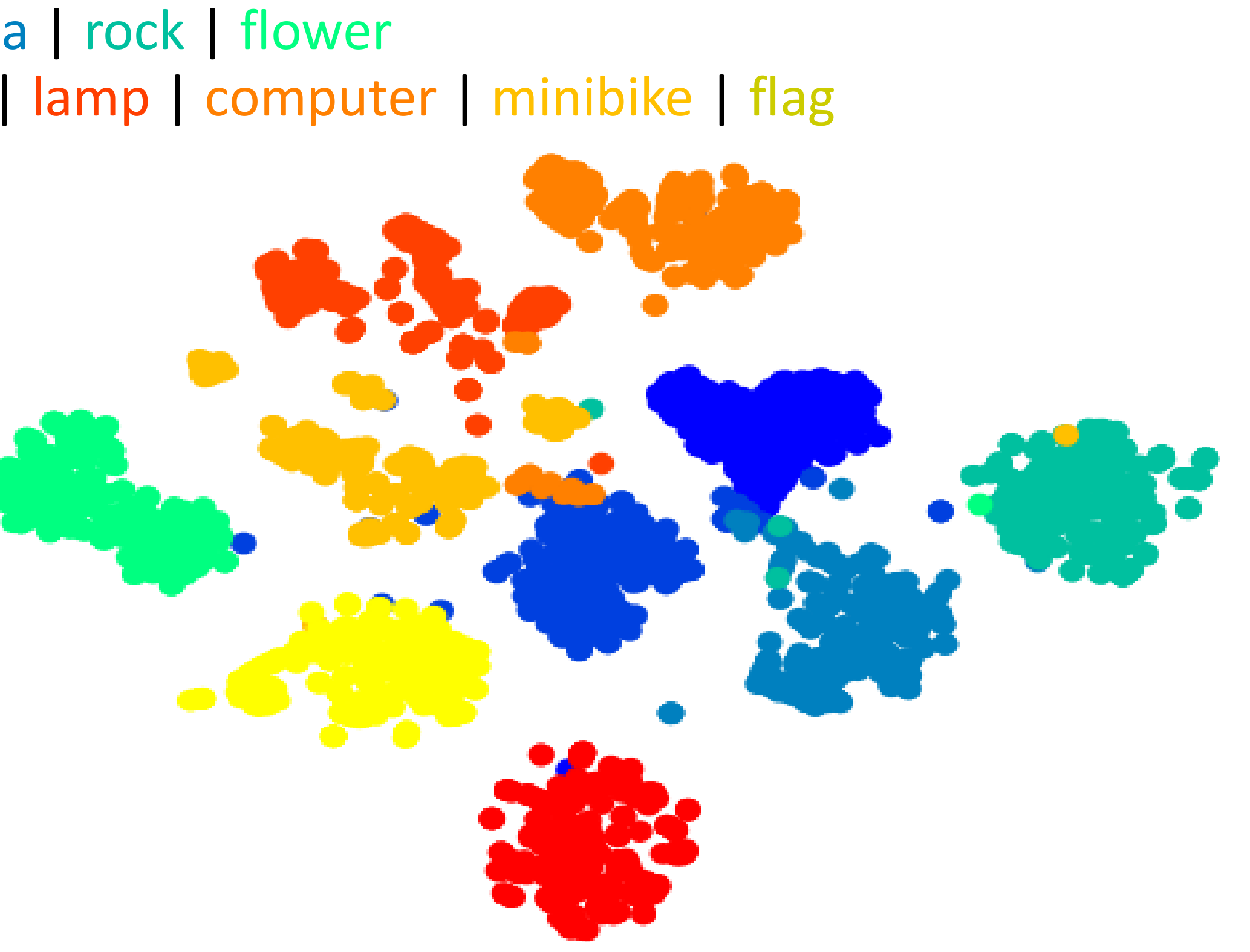}}\hfill\\
        
\vspace{-5pt}
\caption{\textbf{Effects of fine-tuning CLIP.} We show the t-SNE~\cite{van2008visualizing} 
visualization of CLIP image embeddings based on its predictions. In contrast to (a), we observe well-grouped clusters in (b), showing the adaptation of CLIP to segmentation for both seen and unseen classes.} 
\label{fig:embedding_space}
\vspace{-10pt}
\end{figure}

\vspace{-10pt}
\label{finetune}
\paragraph{Analysis on fine-tuning of CLIP.}
In this section, we analyze the effects and methods of fine-tuning of the encoders of CLIP. In Table~\ref{tab:finetuning-ablation}, we report the results of different approaches, which include the variant \textbf{(I)}:~without fine-tuning, \textbf{(II)}:~adopting Prompt Tuning~\cite{zhou2022learning, jia2022visual}, \textbf{(III)}:~fine-tuning the entire CLIP, \textbf{(IV)}:~fine-tuning the attention layer only~\cite{touvron2022three}, \textbf{(V)}:~fine-tuning query and key projections only, \textbf{(VI)}:~fine-tuning key and value projections only, \textbf{(VII)}:~our approach for CLIP image encoder only, \textbf{(VIII)}:~our approach for text encoder only, and  \textbf{(IX)}:~our approach for both encoders. Note that both image and text encoders are fine-tuned in \textbf{(I-VI)}. Overall, we observed that fine-tuning enhances the performance of our framework. Among the various fine-tuning methods, fine-tuning only the query and value projection yields the best performance improvement while also demonstrating high efficiency. Additionally, as can be seen in \textbf{(VII-IX)}, fine-tuning both encoders leads to better performance compared to fine-tuning only one of them in our framework.

In Fig.~\ref{fig:embedding_space}, we show the t-SNE~\cite{van2008visualizing} visualization of the dense image embeddings of CLIP within the A-150~\cite{zhou2019semantic} dataset. We color the embeddings based on the prediction with text classes. From (a), we can observe that the clusters are not well-formed for each classes, due to the image-level training of CLIP. In contrast, we observe well-formed clusters in (b) for both seen and unseen classes, showing the adaptation of CLIP for the downstream task.

\vspace{-10pt}
\paragraph{Training with various datasets.}
In this experiment, we further examine the generalization power of our method in comparison to other methods~\cite{ding2022decoupling, xu2022simple} by training our model on smaller-scale datasets, which include A-150 and PC-59, that poses additional challenges to achieve good performance.  The results are shown in Table~\ref{tab:cross-dataset-ablation}. As shown, we find that although we observe some performance drops, which seem quite natural when a smaller dataset is used, our work significantly outperforms other competitors. These results highlight the strong generalization power of our framework, a favorable characteristic that suggests the practicality of our approach.

\begin{table}[!t]
    \centering
    
    \resizebox{\linewidth}{!}{
    \begin{tabular}{l|c|ccccccc}
    \toprule
        Methods & Training dataset & A-847 & PC-459 & A-150 & PC-59 & PAS-20 & $\textnormal{PAS-20}^b$
        \\
        \midrule\midrule
        ZegFormer & COCO-Stuff & 5.6 & \underline{10.4} & 18.0 & 45.5 & \underline{89.5} & 65.5\\
        ZSseg & COCO-Stuff & \underline{7.0} & 9.0 & \underline{20.5} & \underline{47.7} & 88.4 & \underline{67.9}\\
        \hlrow \ours (ours) & COCO-Stuff & \textbf{12.0} & \textbf{19.0} & \textbf{31.8} & \textbf{57.5} & \textbf{94.6} & \textbf{77.3}\\
        \midrule
        ZegFormer & A-150 & 6.8 & \underline{7.1} & \color{gray}{33.1} & 34.7 & 77.2 & 53.6 \\
        ZSseg & A-150 & \underline{7.6} & \underline{7.1} & \color{gray}{40.3} & \underline{39.7} & \underline{80.9} & \underline{61.1}\\
        \hlrow \ours (ours) & A-150 & \textbf{14.4} & \textbf{16.2} & \color{gray}{47.7} & \textbf{49.9} & \textbf{91.1} & \textbf{73.4} \\
        \midrule
        ZegFormer & PC-59 & \underline{3.8} & \underline{8.2} & \underline{13.1} & \color{gray}{48.7} & 86.5 & 66.8 \\
        ZSseg & PC-59 & 3.0 & 7.6 & 11.9 & \color{gray}{54.7} & \underline{87.7} & \underline{71.7}\\
        \hlrow \ours (ours) & PC-59 & \textbf{9.6} & \textbf{16.7} & \textbf{27.4} & \color{gray}{63.7} & \textbf{93.5} & \textbf{79.9} \\
        \bottomrule
    \end{tabular}
    }

    \vspace{-5pt}
    \caption{\textbf{Training on various datasets.} CLIP with ViT-B is used for all methods. Our model demonstrates remarkable generalization capabilities even on relatively smaller datasets. The scores evaluated on the same dataset used for training are colored in \textcolor{gray}{gray}.}
    \vspace{-10pt}
    \label{tab:cross-dataset-ablation}
\end{table}

\begin{table}[t]
\centering
\resizebox{\linewidth}{!}{
\begin{tabular}{l|ccc|c}
    \toprule
    Methods & ZegFormer & ZSSeg & OVSeg & \ours (Ours) \\
    \midrule\midrule
    \# of learnable params. (M)  & 103.3 & \underline{102.8}& 408.9 & \textbf{70.3}\\
    \# of total params. (M)  & 531.2 & \underline{530.8}& 532.6 & \textbf{433.7}\\
    Training time (min) & 1,148.3& \underline{958.5}&- & \textbf{875.5}\\
    Inference time (s) & 2.70& 2.73&  \underline{2.00}& \textbf{0.54}\\
    Inference GFLOPs &19,425.6 & 22,302.1& \underline{19,345.6}& \textbf{2,121.1}\\
    \bottomrule
\end{tabular}
}
\vspace{-5pt}
\caption{\textbf{Efficiency comparison.} All results are measured with a single RTX 3090 GPU.}
\label{tab:efficiency}
    \vspace{-10pt}
\end{table}

\vspace{-10pt}
\paragraph{Efficiency comparison.}
In Table~\ref{tab:efficiency}, we thoroughly compare the efficiency of our method to recent methods~\cite{ding2022decoupling,xu2022simple,liang2022open}. We measure the number of learnable parameters, the total number of parameters, training time, inference time, and inference GFLOPs. Our model demonstrates strong efficiency in terms of both training and inference. This efficiency is achieved because our framework does not require an additional mask generator~\cite{ding2022decoupling}.

\section{Conclusion}
In conclusion, we introduce a cost aggregation framework for open-vocabulary semantic segmentation, aggregating the cosine-similarity scores between image and text embeddings of CLIP. Through our \ours framework, we fine-tune the encoders of CLIP for its adaptation for the downstream task of segmentation. Our method surpasses the previous state-of-the-art in standard benchmarks and also in scenarios with a vast domain difference. The success in diverse domains underscores the promise and potential of our cost aggregation framework in advancing the field of open-vocabulary semantic segmentation.\\
\vspace{-10pt}\paragraph{Acknowledgement.} This research was supported by the MSIT, Korea (IITP-2023-2020-0-01819, RS-2023-00266509).

\begin{figure*}[t]
    \centering
    \Large \textbf{CAT-Seg: Cost Aggregation for Open-Vocabulary Semantic Segmentation}\\[5pt]
    \large \textbf{-- Supplementary Material --}

    \hsize=1.0\textwidth
    \includegraphics[width=\textwidth]{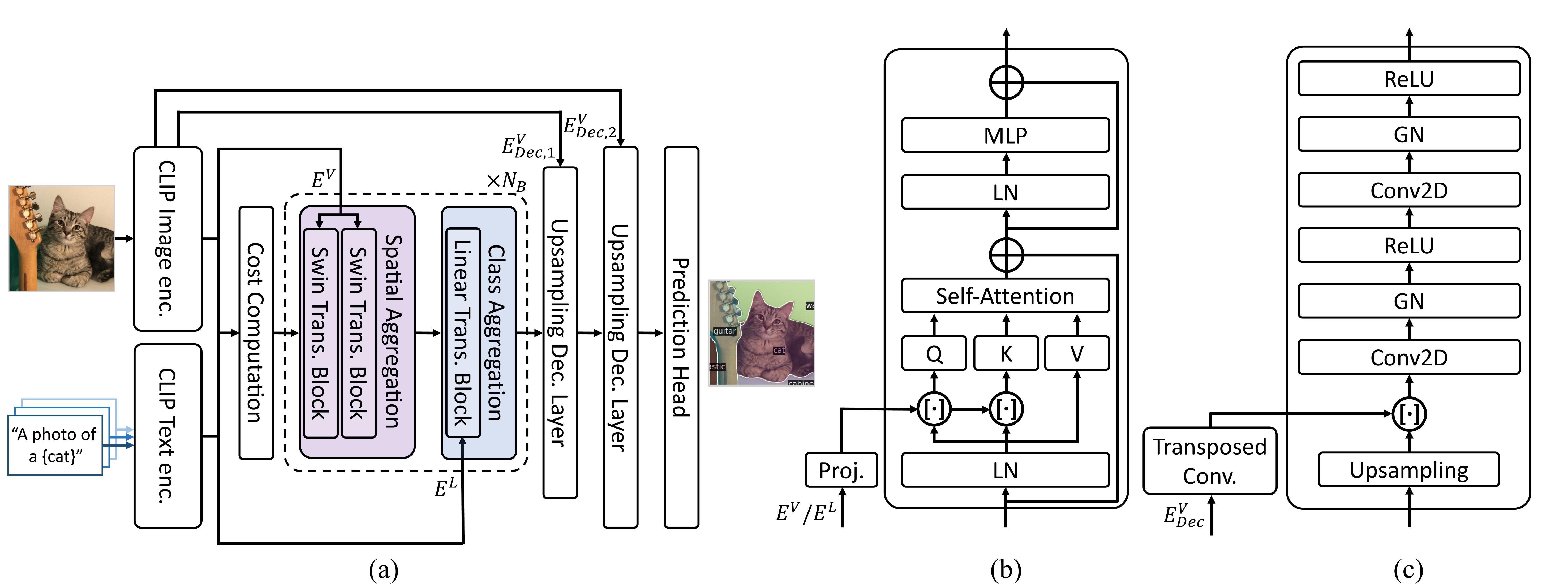}
    \vspace{-10pt}
    \caption{\textbf{More architectural details of CAT-Seg:} (a) overall architecture. (b) embedding guidance. Note that a generalized embedding guidance is illustrated to include different attention designs, \ie, shifted window attention~\cite{liu2021swin} or linear attention~\cite{katharopoulos2020transformers}. (c) upsampling decoder layer. GN: Group Normalization~\cite{wu2018group}. LN: Layer Normalization~\cite{ba2016layer}.}
    \label{fig:guidance-architecture}
    \vspace{-10pt}
\end{figure*}

\renewcommand{\thesection}{\Alph{section}}
\setcounter{section}{0}

In the following, we provide the full results from MESS~\cite{blumenstiel2023mess} in Section~A. We further provide implementation details in Section~B. We then provide additional experimental results and ablation study in Section~C. Finally, we present qualitative results for the benchmarks in Section~D and a discussion of limitations in Section~E.

\section{More Results}\label{A}
\begin{table*}[h]
    \begin{center}

    \resizebox{\textwidth}{!}{
    \begin{tabular}{l|cccccc|ccccc|cccc|cccc|ccc|c}
    \toprule
 & \multicolumn{6}{c|}{General} & \multicolumn{5}{c|}{Earth Monitoring} & \multicolumn{4}{c|}{Medical Sciences} & \multicolumn{4}{c|}{Engineering} & \multicolumn{3}{c|}{Agri. and Biology} & \\
 & \rotatebox[origin=l]{90}{BDD100K} & \rotatebox[origin=l]{90}{Dark Zurich} & \rotatebox[origin=l]{90}{MHP v1} & \rotatebox[origin=l]{90}{FoodSeg103} & \rotatebox[origin=l]{90}{ATLANTIS} & \rotatebox[origin=l]{90}{DRAM} & \rotatebox[origin=l]{90}{iSAID} & \rotatebox[origin=l]{90}{ISPRS Pots.} & \rotatebox[origin=l]{90}{WorldFloods} & \rotatebox[origin=l]{90}{FloodNet} & \rotatebox[origin=l]{90}{UAVid} & \rotatebox[origin=l]{90}{Kvasir-Inst.} & \rotatebox[origin=l]{90}{CHASE DB1} & \rotatebox[origin=l]{90}{CryoNuSeg} & \rotatebox[origin=l]{90}{PAXRay-4} & \rotatebox[origin=l]{90}{Corrosion CS} & \rotatebox[origin=l]{90}{DeepCrack} & \rotatebox[origin=l]{90}{PST900} & \rotatebox[origin=l]{90}{ZeroWaste-f} & \rotatebox[origin=l]{90}{SUIM} & \rotatebox[origin=l]{90}{CUB-200} & \rotatebox[origin=l]{90}{CWFID} & \rotatebox[origin=l]{90}{Mean} \\
\midrule\midrule
\textit{Random (LB)} & \phantom{0}\textit{1.48} & \phantom{0}\textit{1.31} & \phantom{0}\textit{1.27} & \phantom{0}\textit{0.23} & \phantom{0}\textit{0.56} & \phantom{0}\textit{2.16} & \phantom{0}\textit{0.56} & \phantom{0}\textit{8.02} & \textit{18.43} & \phantom{0}\textit{3.39} & \phantom{0}\textit{5.18} & \textit{27.99} & \textit{27.25} & \textit{31.25} & \textit{31.53} & \textit{9.3} & \textit{26.52} & \phantom{0}\textit{4.52} & \phantom{0}\textit{6.49} & \phantom{0}\textit{5.3}\phantom{0} & \phantom{0}\textit{0.06} & \textit{13.08} & \textit{10.27} \\
\textit{Best sup. (UB)} & \textit{44.8}\phantom{0} & \textit{63.9}\phantom{0} & \textit{50.0}\phantom{0} & \textit{45.1}\phantom{0} & \textit{42.22} & \textit{45.71} & \textit{65.3}\phantom{0} & \textit{87.56} & \textit{92.71} & \textit{82.22} & \textit{67.8}\phantom{0} & \textit{93.7}\phantom{0} & \textit{97.05} & \textit{73.45} & \textit{93.77} & \textit{49.92} & \textit{85.9}\phantom{0} & \textit{82.3}\phantom{0} & \textit{52.5}\phantom{0} & \textit{74.0}\phantom{0} & \textit{84.6}\phantom{0} & \textit{87.23} & \textit{70.99} \\
\midrule
ZSSeg-B & 32.36 & 16.86 & \phantom{0}7.08 & \phantom{0}8.17 & 22.19 & 33.19 & \phantom{0}3.8\phantom{0} & 11.57 & 23.25 & 20.98 & 30.27 & 46.93 & \underline{37.0}\phantom{0} & \textbf{38.7} & \underline{44.66} & \phantom{0}3.06 & 25.39 & 18.76 & \phantom{0}8.78 & \underline{30.16} & \phantom{0}4.35 & 32.46 & 22.73 \\
ZegFormer-B & 14.14 & \phantom{0}4.52 & \phantom{0}4.33 & 10.01 & 18.98 & 29.45 & \phantom{0}2.68 & 14.04 & 25.93 & 22.74 & 20.84 & 27.39 & 12.47 & 11.94 & 18.09 & \phantom{0}4.78 & 29.77 & 19.63 & \underline{17.52} & 28.28 & \underline{16.8}\phantom{0} & 32.26 & 17.57 \\
X-Decoder-T & \underline{47.29} & 24.16 & \phantom{0}3.54 & \phantom{0}2.61 & 27.51 & 26.95 & \phantom{0}2.43 & 31.47 & 26.23 & \phantom{0}8.83 & 25.65 & 55.77 & 10.16 & 11.94 & 15.23 & \phantom{0}1.72 & 24.65 & 19.44 & 15.44 & 24.75 & \phantom{0}0.51 & 29.25 & 19.8\phantom{0} \\
SAN-B & 37.4\phantom{0} & 24.35 & \phantom{0}8.87 & \underline{19.27} & \underline{36.51} & 49.68 & \phantom{0}4.77 & \underline{37.56} & 31.75 & \underline{37.44} & \textbf{41.65} & \underline{69.88} & 17.85 & 11.95 & 19.73 & \phantom{0}3.13 & \underline{50.27} & 19.67 & \textbf{21.27} & 22.64 & \textbf{16.91} & \phantom{0}5.67 & 26.74 \\
OpenSeeD-T & \textbf{47.95} & \underline{28.13} & \phantom{0}2.06 & \phantom{0}9.0\phantom{0} & 18.55 & 29.23 & \phantom{0}1.45 & 31.07 & 30.11 & 23.14 & 39.78 & 59.69 & \textbf{46.68} & 33.76 & 37.64 & 13.38 & 47.84 & \phantom{0}2.5\phantom{0} & \phantom{0}2.28 & 19.45 & \phantom{0}0.13 & 11.47 & 24.33 \\
Gr.-SAM-B & 41.58 & 20.91 & \textbf{29.38} & 10.48 & 17.33 & \underline{57.38} & \underline{12.22} & 26.68 & \underline{33.41} & 19.19 & 38.34 & 46.82 & 23.56 & \underline{38.06} & 41.07 & \textbf{20.88} & \textbf{59.02} & \textbf{21.39} & 16.74 & 14.13 & \phantom{0}0.43 & \underline{38.41} & \underline{28.52} \\
\hlrow CAT-Seg-B & 46.71 &\textbf{28.86} &\underline{23.74} &\textbf{26.69} &\textbf{40.31} &\textbf{65.81} &\textbf{19.34} &\textbf{45.36} &\textbf{35.72} &\textbf{37.57} &\underline{41.55} &48.2\phantom{0} &16.99 &15.7\phantom{0} &31.48 &12.29 &31.67 &\underline{19.88} &\underline{17.52} &\textbf{44.71} &10.23 &\textbf{42.77} & \textbf{31.96} \\

\midrule
OVSeg-L & \underline{45.28} & 22.53 & \phantom{0}6.24 & 16.43 & 33.44 & 53.33 & \phantom{0}8.28 & 31.03 & 31.48 & 35.59 & 38.8 & \underline{71.13} & \underline{20.95} & \underline{13.45} & 22.06 & \phantom{0}6.82 & 16.22 & \underline{21.89} & 11.71 & 38.17 & 14.0\phantom{0} & 33.76 & 26.94 \\
SAN-L & 43.81 & \underline{30.39} & \phantom{0}9.34 & \underline{24.46} & \underline{40.66} & \underline{68.44} & 11.77 & \textbf{51.45} & \textbf{48.24} & \underline{39.26} & \textbf{43.41} & \textbf{72.18} & \phantom{0}7.64 & 11.94 & \underline{29.33} & \phantom{0}6.83 & \underline{23.65} & 19.01 & \underline{18.32} & \underline{40.01} & \underline{19.3}\phantom{0} & \phantom{0}1.91 & \underline{30.06} \\
Gr.-SAM-L & 42.69 & 21.92 & \underline{28.11} & 10.76 & 17.63 & 60.8\phantom{0} & \underline{12.38} & 27.76 & 33.4\phantom{0} & 19.28 & 39.37 & 47.32 & \textbf{25.16} & \textbf{38.06} & \textbf{44.22} & \textbf{20.88} & \textbf{58.21} & 21.23 & 16.67 & 14.3\phantom{0} & \phantom{0}0.43 & \underline{38.47} & 29.05 \\
\hlrow CAT-Seg-L & \textbf{47.87} &\textbf{34.96} &\textbf{32.54} &\textbf{33.31} &\textbf{45.61} &\textbf{73.82} & \textbf{20.58} & \underline{50.81} & \underline{46.42} & \textbf{41.36} & \underline{40.79} &61.13 &3.72 &11.94 &22.02 &\underline{11.03} &19.9\phantom{0} &\textbf{22.0}\phantom{0} &\textbf{27.87} &\textbf{53.0}\phantom{0} &\textbf{22.93} &\textbf{39.91} &\textbf{34.7}\phantom{0}\\
        \bottomrule
    \end{tabular}
    }
    \end{center}
    \vspace{-15pt}
    \caption{\textbf{Full results of quantitative evaluation on MESS~\citep{blumenstiel2023mess}.} 
    }\label{tab:mess_full}
    \vspace{-10pt}
\end{table*}

\paragraph{Full quantitative results on MESS benchmark.}
In Table~\ref{tab:mess_full}, we provide the results of all 22 datasets within MESS~\cite{blumenstiel2023mess}, including results from Grounded-SAM~\cite{Grounded-SAM_Contributors_Grounded-Segment-Anything_2023}.

\section{More Details}\label{B}
\subsection{Architectural Details}

In the following, we provide more architectural details. Our detailed overall architecture is illustrated in Fig.~\ref{fig:guidance-architecture} (a).

\smallbreak
\noindent\textbf{Embedding guidance.}
In this paragraph, we provide more details of embedding guidance, which is designed to facilitate the cost aggregation process by exploiting its rich semantics for a guidance.  We first extract visual and text embeddings from CLIP encoders~\cite{radford2021learning}. The embeddings then undergo linear projection and concatenated to the cost volume before query and key projections in aggregation layer. The design is illustrated in Fig.~\ref{fig:guidance-architecture}~(b).

\smallbreak
\noindent\textbf{Upsampling decoder.}
The detailed architecture is illustrated in Fig.~\ref{fig:guidance-architecture}(c). In our upsampling decoder, we start by taking high-resolution features from the CLIP ViT model~\cite{dosovitskiy2020image}. We then apply a single transposed convolution layer to these extracted features to generate an upsampled feature map. Initially, the extracted feature maps have a resolution of $24\times 24$ pixels. However, after processing them with the transposed convolution operation, we increase their resolution to $48\times 48$ pixels for the first feature map, denoted as $E^V_{Dec,1}$, and to $96\times 96$ pixels for the second feature map, denoted as $E^V_{Dec,2}$.

To obtain $E^V_{Dec,1}$, we utilize the output of the 8th layer for the ViT-B/16 model, and for the ViT-L/14 model, we use the output of the 16th layer. For the extraction of $E^V_{Dec,2}$, we employ shallower features: the output of the 4th layer for the ViT-B/16 model as a VLM, and the output of the 8th layer for the ViT-L/14 model. These features are employed to enhance cost embeddings with fine details using a U-Net-like architecture~\cite{ronneberger2015u}.

\subsection{Other Implementation Details}
\smallbreak
\noindent\textbf{Training details.}
A resolution of $H=W=24$ is used during training for constructing cost volume. The position embeddings of the CLIP image encoder is initialized with bicubic interpolation~\cite{touvron2021training}, and we set training resolution as $384\times 384$. For ViT-B and ViT-L variants, we initialize CLIP~\cite{radford2021learning} with official weights of ViT-B/16 and ViT-L/14@336px respectively. All hyperparameters are kept constant across the evaluation datasets.  

\smallbreak
\noindent\textbf{Text prompt templates.}
To obtain text embeddings from the text encoder, we form sentences with the class names, such as \texttt{"A photo of a \{class\}"}. We do not explore handcrafted prompts in this work, but it is open for future investigation. %

\begin{table*}[h]
    \begin{center}
       
    \resizebox{\textwidth}{!}{
    \begin{tabular}{lllccl}
\toprule
Dataset & Link & Licence & Split & \# of classes & Classes \\
\midrule\midrule
BDD100K \cite{yu2020bdd100k} & \href{https://bdd-data.berkeley.edu}{berkeley.edu} & \href{https://doc.bdd100k.com/license.html}{custom} & val & 19 & [road; sidewalk; building; wall; fence; pole; traffic light; traffic sign; ...] \\
Dark Zurich \cite{sakaridis2019guided} & \href{https://www.trace.ethz.ch/publications/2019/GCMA_UIoU/}{ethz.ch} & custom & val & 20 & [unlabeled; road; sidewalk; building; wall; fence; pole; traffic light; ...] \\
MHP v1 \cite{li2017multiple} & \href{https://github.com/ZhaoJ9014/Multi-Human-Parsing}{github.com} & \href{https://lv-mhp.github.io/}{custom} & test & 19 & [others; hat; hair; sunglasses; upper clothes; skirt; pants; dress; ...] \\
FoodSeg103 \cite{wu2021large} & \href{https://xiongweiwu.github.io/foodseg103.html}{github.io} & Apache 2.0 & test & 104 & [background; candy; egg tart; french fries; chocolate; biscuit; popcorn; ...] \\
ATLANTIS \cite{erfani2022atlantis} & \href{https://github.com/smhassanerfani/atlantis}{github.com} & Flickr (images) & test & 56 & [bicycle; boat; breakwater; bridge; building; bus; canal; car; ...] \\
DRAM \cite{cohen2022semantic} & \href{https://faculty.runi.ac.il/arik/site/artseg/Dram-Dataset.html}{ac.il} & custom (in download) & test & 12 & [bird; boat; bottle; cat; chair; cow; dog; horse; ...] \\
\hline
iSAID \cite{waqas2019isaid} & \href{https://captain-whu.github.io/iSAID/dataset.html}{github.io} & Google Earth (images) & val & 16 & [others; boat; storage tank; baseball diamond; tennis court; bridge; ...] \\
ISPRS Potsdam \cite{DatasetISPRSPotsdam} & \href{https://www.isprs.org/education/benchmarks/UrbanSemLab/default.aspx}{isprs.org} & no licence provided\footnote{Upon request, the naming of the data provider and project is required.} & test & 6 & [road; building; grass; tree; car; others] \\
WorldFloods \cite{mateo2021towards} & \href{https://github.com/spaceml-org/ml4floods/blob/main/jupyterbook/content/worldfloods_dataset.md}{github.com} & CC NC 4.0 & test & 3 & [land; water and flood; cloud] \\
FloodNet \cite{rahnemoonfar2021floodnet} & \href{https://github.com/BinaLab/FloodNet-Supervised_v1.0}{github.com} & \href{https://cdla.dev/permissive-1-0/}{custom} & test & 10 & [building-flooded; building-non-flooded; road-flooded; water; tree; ...] \\
UAVid \cite{lyu2020uavid} & \href{https://uavid.nl}{uavid.nl} & CC BY-NC-SA 4.0 & val & 8 & [others; building; road; tree; grass; moving car; parked car; humans] \\
\hline
Kvasir-Inst. \cite{jha2021kvasir} & \href{https://datasets.simula.no/kvasir-instrument/}{simula.no} & \href{https://datasets.simula.no/kvasir-instrument/}{custom} & test & 2 & [others; tool] \\
CHASE DB1 \cite{fraz2012ensemble} & \href{https://blogs.kingston.ac.uk/retinal/chasedb1/}{kingston.ac.uk} & CC BY 4.0 & test & 2 & [others; blood vessels] \\
CryoNuSeg \cite{mahbod2021cryonuseg} & \href{https://www.kaggle.com/datasets/ipateam/segmentation-of-nuclei-in-cryosectioned-he-images}{kaggle.com} & CC BY-NC-SA 4.0 & test & 2 & [others; nuclei in cells] \\
PAXRay-4 \cite{seibold2022detailed} & \href{https://constantinseibold.github.io/paxray/}{github.io} & \href{https://constantinseibold.github.io/paxray/}{custom} & test & 4x2 & [others, lungs], [others, bones], [others, mediastinum], [others, diaphragm] \\
\hline
Corrosion CS \cite{bianchi2021corrosion} & \href{https://figshare.com/articles/dataset/Corrosion_Condition_State_Semantic_Segmentation_Dataset/16624663}{figshare.com} & CC0 & test & 4 & [others; steel with fair corrosion; ... poor corrosion; ... severe corrosion] \\
DeepCrack \cite{liu2019deepcrack} & \href{https://github.com/yhlleo/DeepCrack/tree/master}{github.com} & \href{https://github.com/yhlleo/DeepCrack/tree/master}{custom} & test & 2 & [concrete or asphalt; crack] \\
PST900 \cite{shivakumar2020pst900} & \href{https://github.com/ShreyasSkandanS/pst900_thermal_rgb}{github.com} & GPL-3.0 & test & 5 & [background; fire extinguisher; backpack; drill; human] \\
ZeroWaste-f \cite{bashkirova2022zerowaste} & \href{http://ai.bu.edu/zerowaste/}{ai.bu.edu} & CC-BY-NC 4.0 & test & 5 & [background or trash; rigid plastic; cardboard; metal; soft plastic] \\
\hline
SUIM \cite{islam2020semantic} & \href{https://irvlab.cs.umn.edu/resources/suim-dataset}{umn.edu} & MIT & test & 8 & [human diver; reefs and invertebrates; fish and vertebrates; ...] \\
CUB-200 \cite{welinder2010caltech} & \href{https://www.vision.caltech.edu/datasets/cub_200_2011/}{caltech.edu} & \href{https://www.vision.caltech.edu/datasets/cub_200_2011/}{custom} & test & 201 & [background; Laysan Albatross; Sooty Albatross; Crested Auklet; ...] \\
CWFID \cite{haug2015crop} & \href{https://github.com/cwfid/dataset}{github.com} & \href{https://github.com/cwfid/dataset}{custom} & test & 3 & [ground; crop seedling; weed] \\
\bottomrule
    \end{tabular}
}
    \end{center}
    \vspace{-15pt}
        \caption{\textbf{Details of the datasets in the MESS benchmark~\cite{blumenstiel2023mess}.} 
    }\label{tab:mess-detail}
    \vspace{-10pt}
\end{table*}

\begin{figure}[t]
    \centering
    \includegraphics[width=1.0\linewidth]{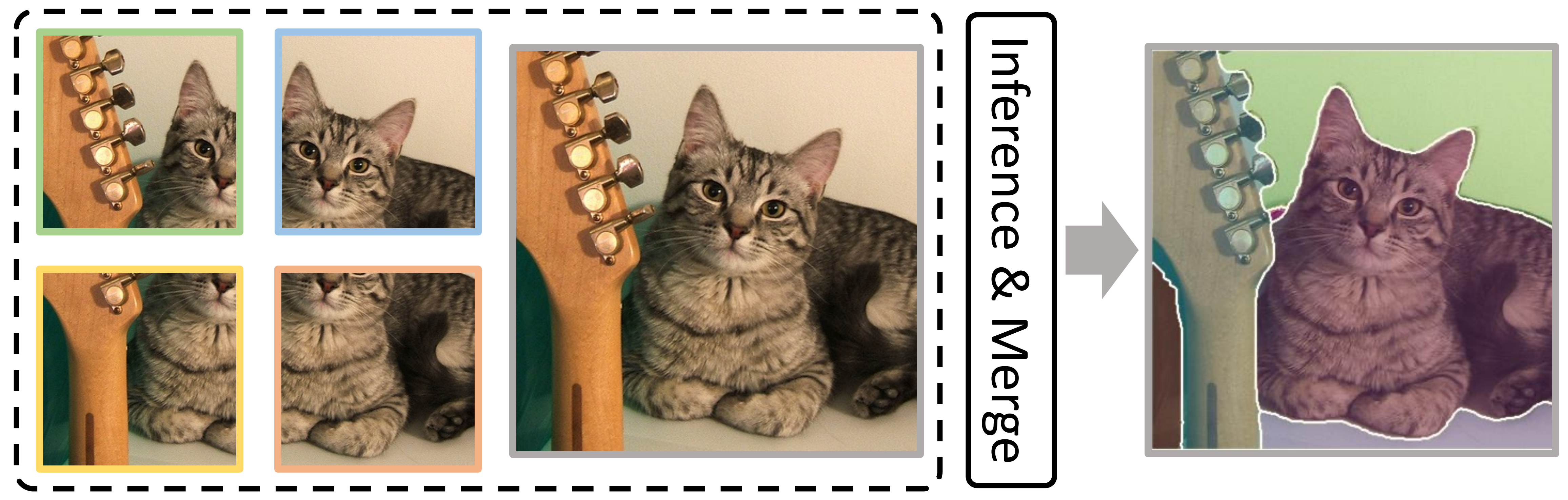}\hfill\\
    \vspace{-5pt}
    \caption{\textbf{Illustration of the patch inference.} During inference, we divide the input image into patches, thereby increasing the effective resolution.}
    \label{fig:patch-inference}
    \vspace{-10pt}
\end{figure}

\subsection{Patch Inference}
The practicality of Vision Transformer (ViT)~\cite{dosovitskiy2020image} for high-resolution image processing has been limited due to its quadratic complexity with respect to the sequence length. As our model leverages ViT to extract image embeddings, \ours may struggle to output to the conventional image resolutions commonly employed in semantic segmentation literature, such as $640\times 640$~\cite{cheng2021per,ghiasi2022scaling}, without sacrificing some accuracy made by losing some fine-details. Although we can adopt the same approach proposed in~\cite{zhou2022extract} to upsample the positional embedding~\cite{zhou2022extract}, we ought to avoid introducing excessive computational burdens, and thus adopt an effective inference strategy without requiring additional training which is illustrated in Fig.~\ref{fig:patch-inference}.

To this end, we begin by partitioning the input image into overlapping patches of size $\frac{H}{N_P} \times \frac{W}{N_P}$. Intuitively, given an image size of $640 \times 640$, we partition the image to sub-images of size $384 \times 384$, which matches the image resolution at training phase, and each sub-images has overlapping regions $128 \times 128$. Subsequently, we feed these sub-images and the original image that is resized to $384 \times 384$ into the model. Given the results for each patches and the image, we merge the obtained prediction, while the overlapping regions are averaged to obtain the final prediction. In practice, we employ $N_P=2$,  while adjusting the overlapping region to match the effective resolution of $640\times 640$.

\subsection{More Details of MESS Benchmark}

In Table~\ref{tab:mess-detail}, we provide details of the datasets in the MESS benchmark~\cite{blumenstiel2023mess}.

\section{Additional Ablation Study}\label{C}

\subsection{Ablation Study of Inference Strategy}\vspace{-10pt}
\begin{table}[H]
    \centering
    \resizebox{\linewidth}{!}{
   \begin{tabular}{l|cccccc}
        \toprule
        Methods & A-847 & PC-459 & A-150 & PC-59 & PAS-20 & $\textnormal{PAS-20}^b$
         \\
        \midrule\midrule
        \ours w/ training reso. & \underline{14.6}& \underline{22.1}& \underline{35.7}& \underline{60.9}& \underline{96.3}& \underline{79.9}\\
        \hlrow Ours & \textbf{16.0}& \textbf{23.8}& \textbf{37.9}& \textbf{63.3}& \textbf{97.0}& \textbf{82.5}\\
        \bottomrule
\end{tabular}
}%
\vspace{-5pt}
        \caption{\textbf{Ablation study of inference strategy.} CLIP with ViT-L is used for ablation.}
    \label{tab:inference-ablation}
    \vspace{-10pt}
\end{table}

Table~\ref{tab:inference-ablation} presents effects of different inference strategies for our model. The first row shows the results using the training resolution at inference time. The last row adopts the proposed patch inference strategy. It is shown that our proposed approach can bring large performance gains, compared to using the training resolution.

\subsection{Ablation on VLM}

\begin{table}[H]
\centering
\resizebox{\linewidth}{!}{
\begin{tabular}{l|cccccc}
        \toprule
        VLM & A-847 & PC-459 & A-150 & PC-59 & PAS-20 & $\textnormal{PAS-20}^b$
         \\
        \midrule\midrule
        EVA-02-CLIP-L/14~\cite{sun2023eva} & \underline{16.4} & \underline{24.5} & 37.8 & \underline{62.7} & \textbf{97.9} & \textbf{83.7}\\
        SigLIP-ViT-L/16~\cite{zhai2023sigmoid} & \textbf{18.0}& \textbf{26.1} & \textbf{39.1} & 60.9& \underline{97.2}&80.8\\
        \hlrow CLIP-ViT-L/14 & 16.0 & 23.8 & \underline{37.9} & \textbf{63.3} & 97.0 & \underline{82.5} \\
        \bottomrule
\end{tabular}}
    \vspace{-5pt}
\caption{\textbf{Results on various VLMs.}}
    \vspace{-10pt}
    \label{tab:vlms}
\end{table}

Table~\ref{tab:vlms} shows the results with various VLMs. We found that \ours can be applied to various VLMs, and better results can be obtained when a more powerful model is applied.

\section{More Qualitative Results}\label{D}
\begin{figure*}
    \centering
    \includegraphics[width=0.95\linewidth]{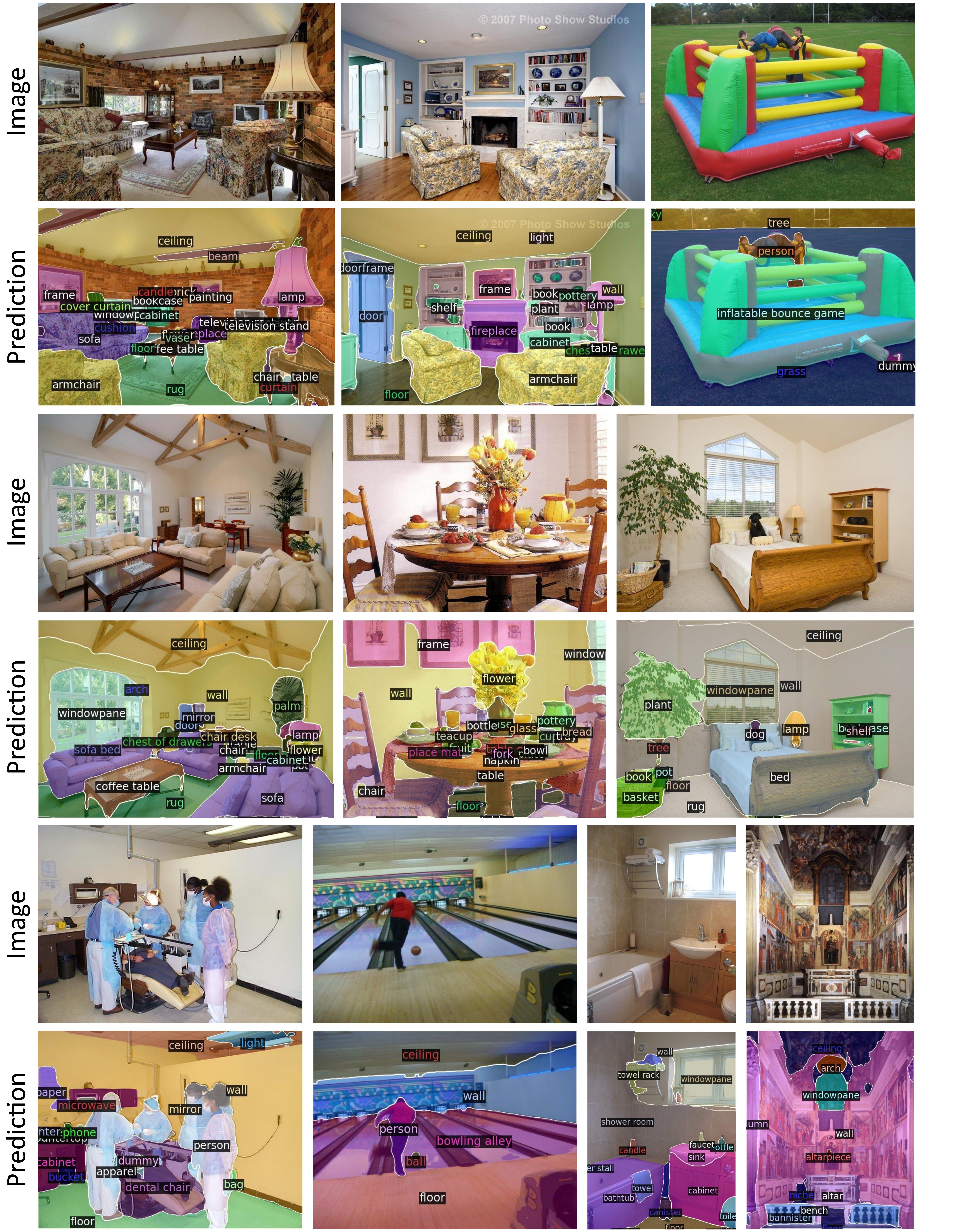}\hfill\\
    \vspace{-5pt}
    \caption{\textbf{Qualitative results on ADE20K~\citep{zhou2019semantic} with 847 categories.}}
    \label{fig:ade847}\vspace{-5pt}
\end{figure*}

\begin{figure*}
    \centering
    \includegraphics[width=0.95\linewidth]{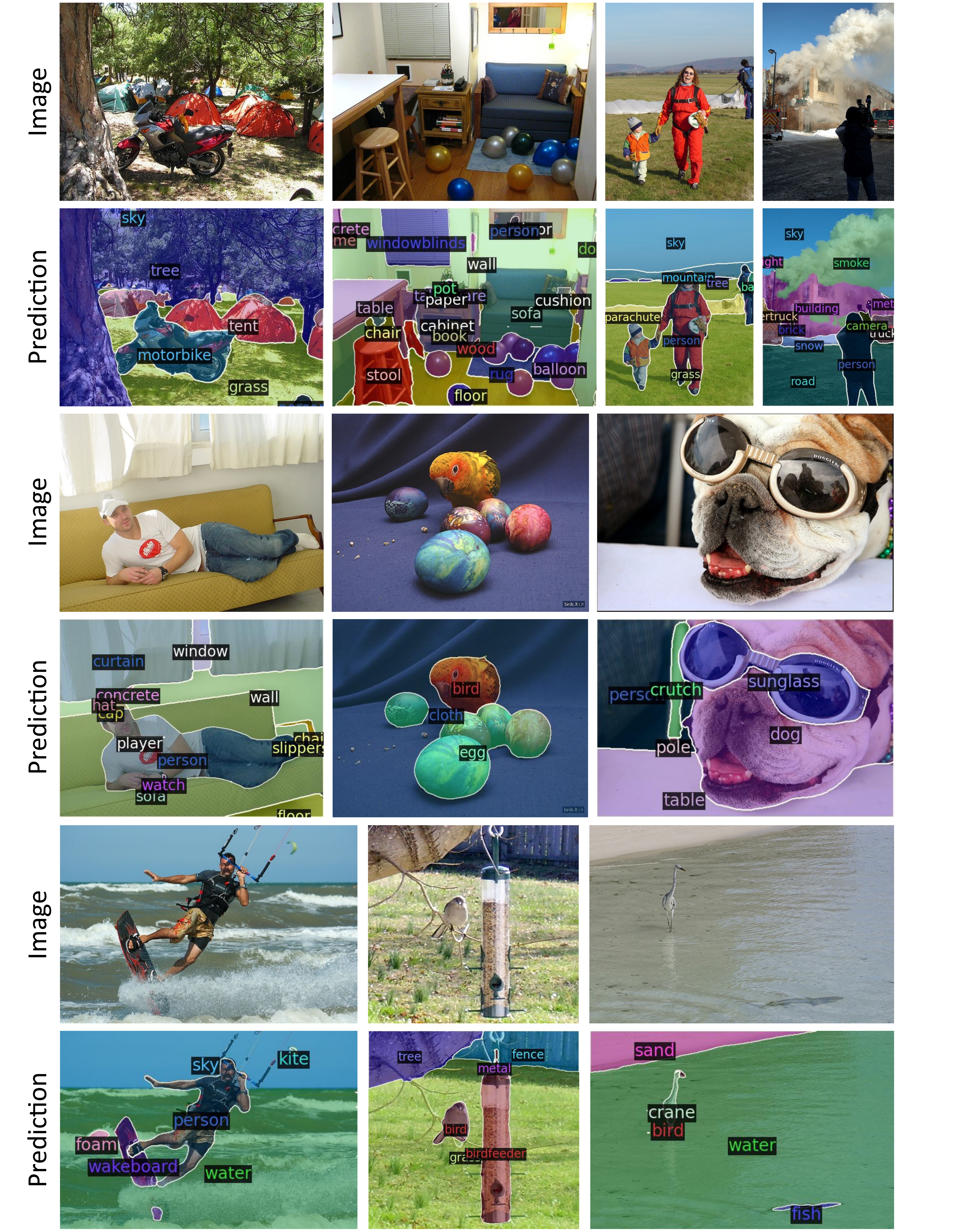}\hfill\\
    \vspace{-5pt}
    \caption{\textbf{Qualitative results on PASCAL Context~\citep{mottaghi2014role} with  459 categories.}}
    \label{fig:pc459}\vspace{-5pt}
\end{figure*}

\begin{figure*}
    \centering
    \includegraphics[width=0.95\linewidth]{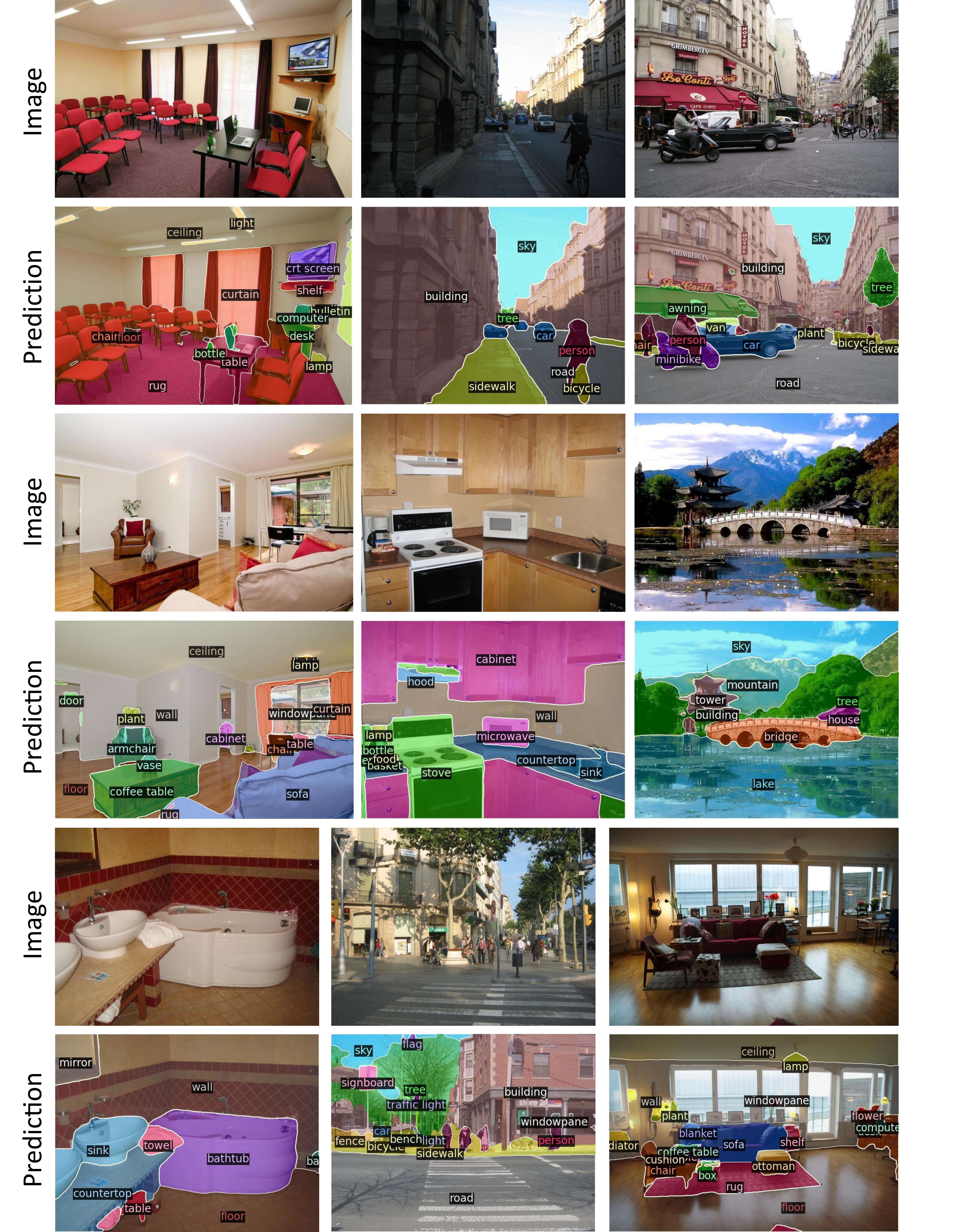}\hfill\\
    \vspace{-5pt}
    \caption{\textbf{Qualitative results on ADE20K~\citep{zhou2019semantic} with 150 categories.}}
    \label{fig:ade150}\vspace{-5pt}
\end{figure*}

\begin{figure*}
    \centering
    \includegraphics[width=0.95\linewidth]{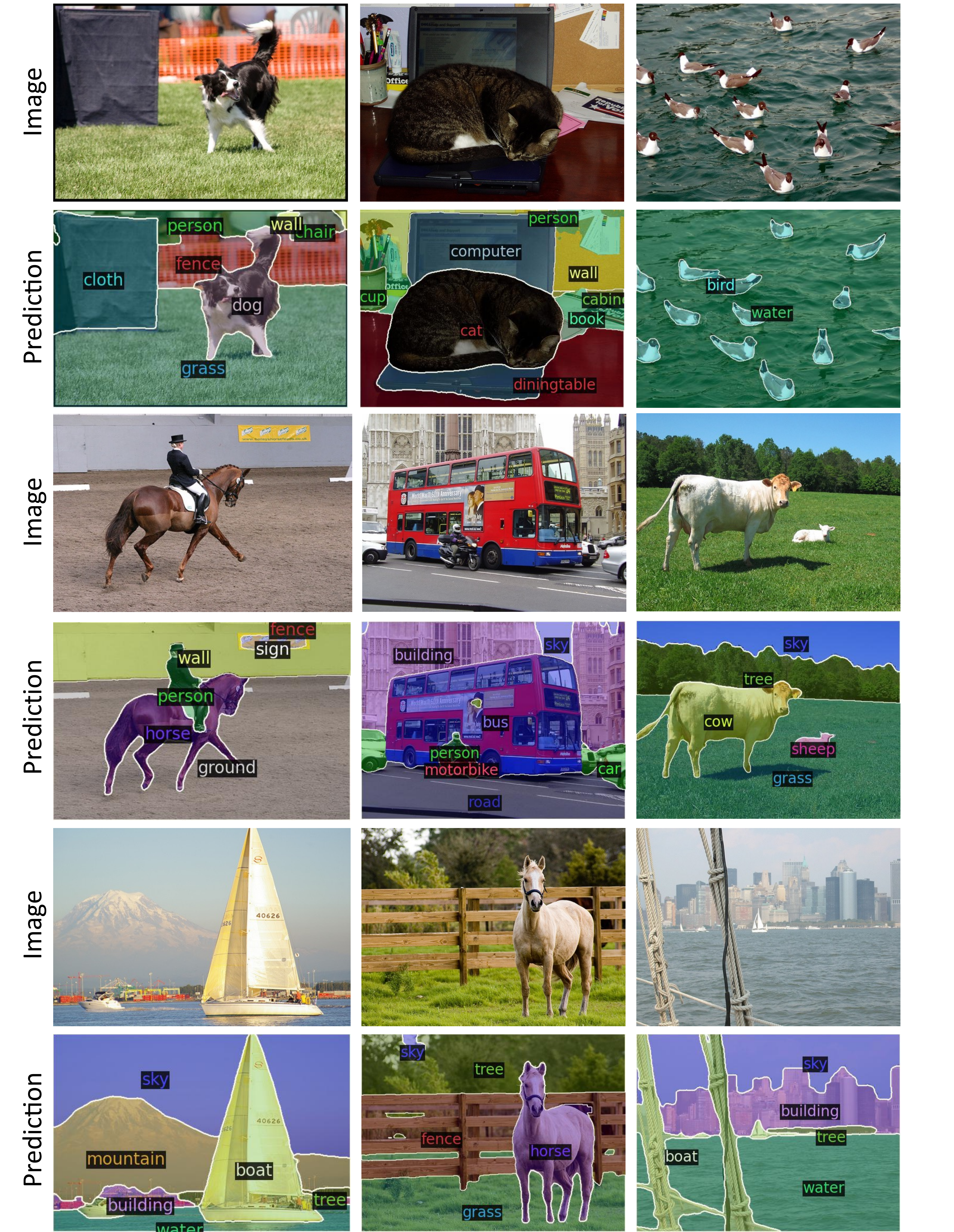}\hfill\\
    \vspace{-5pt}
    \caption{\textbf{Qualitative results on PASCAL Context~\citep{mottaghi2014role} with 59 categories.}}
    \label{fig:pc59}\vspace{-5pt}
\end{figure*}

\begin{figure*}
    \centering
    \includegraphics[width=1.0\linewidth]{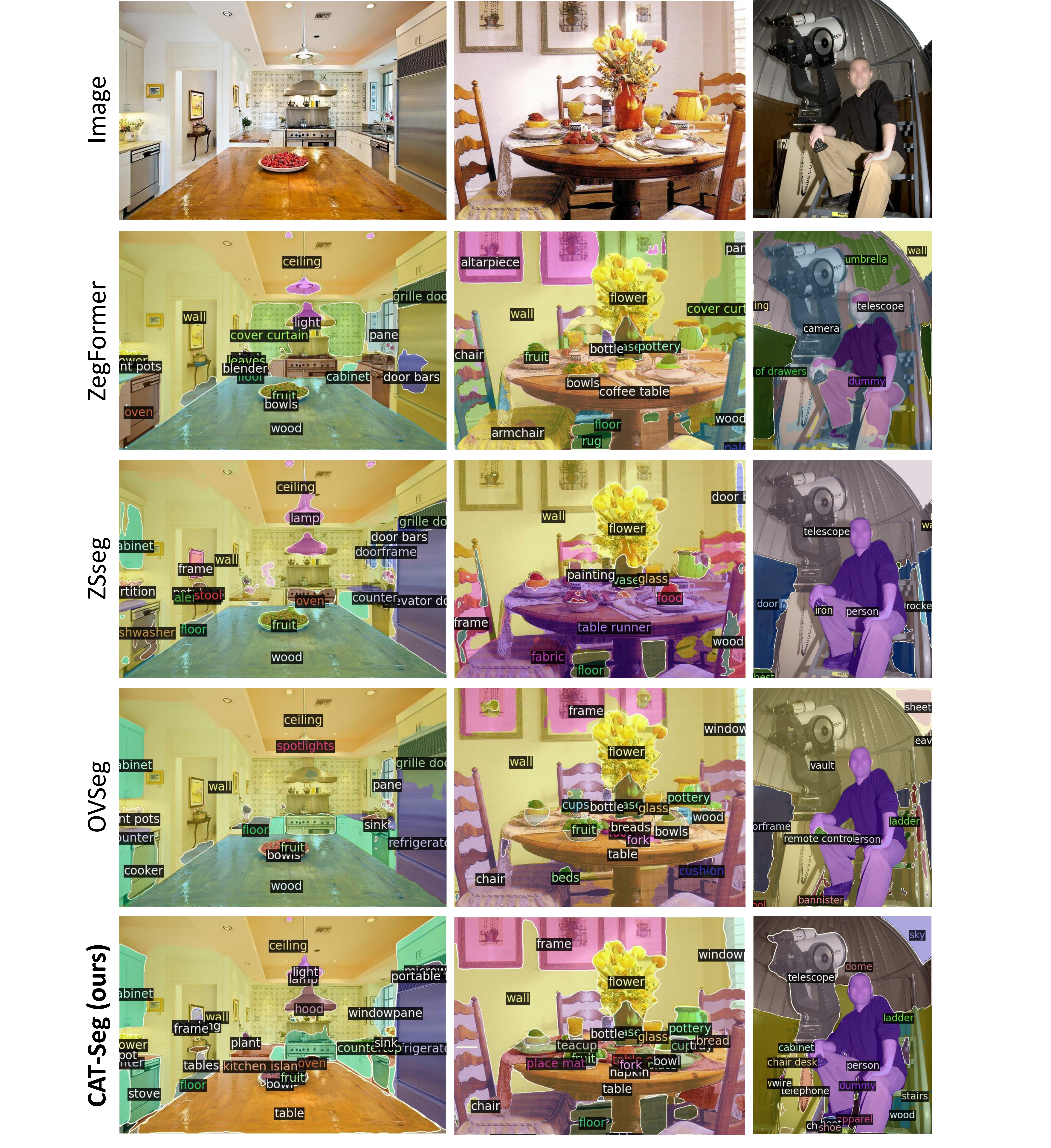}\hfill\\
    \vspace{-5pt}
    \caption{\textbf{Comparison of qualitative results on ADE20K~\citep{zhou2019semantic} with 847 categories.} We compare CAT-Seg with ZegFormer~\citep{ding2022decoupling}, ZSseg~\citep{xu2022simple}, and OVSeg~\citep{liang2022open} on A-847 dataset.}
    \label{fig:ade847-comparison}\vspace{-5pt}
\end{figure*}

We provide more qualitative results on A-847~\cite{zhou2019semantic} in Fig.~\ref{fig:ade847}, PC-459~\cite{mottaghi2014role} in Fig.~\ref{fig:pc459}, A-150~\cite{zhou2019semantic} in Fig.~\ref{fig:ade150}, and PC-59~\cite{mottaghi2014role} in Fig.~\ref{fig:pc59}. We also further compare the results in A-847~\cite{zhou2019semantic} with other methods~\cite{ding2022decoupling, xu2022simple, liang2022open} in Fig.~\ref{fig:ade847-comparison}.

\section{Limitations}\label{E}
To evaluate open-vocabulary semantic segmentation results, we follow ~\cite{ghiasi2022scaling, liang2022open} and compute the metrics using the other segmentation datasets. However, since the ground-truth segmentation maps involve some ambiguities, the reliability of  the evaluation dataset is somewhat questionable. %
Constructing a more reliable dataset including ground-truths accounting for above issue for accurate evaluation is an intriguing topic.

\clearpage
\clearpage

{
    \small
    \bibliographystyle{ieeenat_fullname}
    \bibliography{main}
}
\clearpage

\end{document}